%% file: main.tex
\documentclass{article}
\usepackage{fontspec}
\newfontfamily\cjkfont{FandolSong-Regular.otf}[Script=CJK]
\XeTeXlinebreaklocale "zh"
\usepackage{graphicx,booktabs,longtable,array,calc}
\usepackage{amsmath,amssymb,bm}
\usepackage{float,caption,fvextra,xurl}
\usepackage{hyperref}
\usepackage[preprint]{icml2026}
\renewcommand{\today}{November 12, 2026}
\hypersetup{pdftitle={FPSRG: Heterogeneous User Simulation},pdfauthor={Ruimin Pei, Haibo Liu, Fan Zhang, Xunhua Guo}}
\providecommand{\tightlist}{\setlength{\itemsep}{0pt}\setlength{\parskip}{0pt}}

\newcommand{\pandocbounded}[1]{#1}

\setkeys{Gin}{width=\linewidth,keepaspectratio}
\icmltitlerunning{FPSRG: Heterogeneous User Simulation}
\begin{document}
\twocolumn[
\icmltitle{FPSRG: Heterogeneous User Simulation\\First-Person Post-Training and Multidimensional Semantic Response Grids}
\input{authors.tex}
\icmlkeywords{user simulation, activation steering, joint calibration}
\vskip 0.3in
]
\hypersetup{pdftitle={FPSRG: Heterogeneous User Simulation -- First-Person Post-Training and Multidimensional Semantic Response Grids},pdfsubject={Research preprint}}
\printAffiliationsAndNotice{Email for Ruimin Pei: \texttt{prm21@mails.tsinghua.edu.cn}. }
\begin{abstract}\input{abstract.tex}\end{abstract}
\input{body.tex}
\clearpage

\input{main.bbl}
\clearpage
\onecolumn
\input{appendix.tex}
\end{document}

%% file: authors.tex
\begin{icmlauthorlist}
\icmlauthor{Ruimin Pei}{tsinghua}
\icmlauthor{Haibo Liu}{yoolee}
\icmlauthor{Fan Zhang}{yoolee}
\icmlauthor{Xunhua Guo}{tsinghua}
\end{icmlauthorlist}
\icmlaffiliation{tsinghua}{School of Economics and Management, Tsinghua University, Beijing, China}
\icmlaffiliation{yoolee}{yoolee.ai}
\icmlcorrespondingauthor{Xunhua Guo}{guoxh@sem.tsinghua.edu.cn}

%% file: abstract.tex
Simulating heterogeneous individuals with large language models is an active research direction, and heterogeneous user simulation requires models with different characteristics. We study profile construction: given dimensions, items, and scoring rules, build model configurations that can be retrieved by multidimensional target scores. We introduce First-Person Semantic Response Grids (FPSRG). Shared post-training first teaches the model to judge and answer as the participant; self-report and behavioral targets then refine dimension-specific control vectors. With the model frozen, we measure joint scores for vector-coefficient combinations, filter inadmissible configurations, and store the results in a searchable catalog. Per-token configuration routing, continuous batching, and isolated caches support parallel measurement of different profiles. Non-thinking five- and ten-dimensional Qwen instances and a thinking-mode six-dimensional instance following full-parameter post-training of Huihui demonstrate the complete workflow. With matched data, layers, and norms, refinement increases mean five-axis composite control span from 30.75 to 40.45. Within the five-dimensional 25--75 target region, the qualified five-level vector catalog covers 92.2\% of fixed targets under per-axis error at most 20 and total error at most 60; combining it with the prompt catalog increases coverage to 98.7\%. At a fixed coefficient range, three-, five-, and nine-level catalogs achieve 71.2\%, 92.2\%, and 94.6\% coverage, demonstrating the gains from denser measurement. These results describe measured reachable regions under a fixed assessment protocol. Through shared adaptation, a small set of learned directions, and reusable joint measurements, FPSRG constructs heterogeneous individual configurations without retraining a model for every profile.

%% file: body.tex
\section*{1 Introduction}\label{introduction}

Facing an unfinished task, some people worry about the deadline, others prioritize commitments, and others negotiate responsibilities. Customer-service training, interactive evaluation, and social research need to construct these differences. Language models provide simulated participants \citep{ref1,ref2,ref3} for comparing how different attribute combinations respond to the same information. We study \textbf{heterogeneous model configurations that can be retrieved by target scores and repeatedly measured within a specified multidimensional evaluation system.}

Prompts describe high or low tendencies, but ``very,'' ``moderately,'' and ``slightly'' still require score calibration. Training a LoRA for every profile introduces repeated optimization and storage as combinations increase. Personality Sliders, PERSONA, and ORBIT demonstrate strength adjustment and attribute composition \citep{ref7,ref8,ref9}. Mapping coefficients to semantic coordinates additionally requires handling cross-dimensional responses, saturation, and interaction: single-axis effects do not determine combination outcomes.

First-Person Semantic Response Grids (FPSRG) first use shared post-training to organize expression around a participant's facts, goals, and trade-offs, then refine directions with self-report and behavioral targets. With the model frozen, we measure complete score vectors for coefficient combinations and store a searchable configuration--score catalog. This retains coupling and retrieves by target error. Prompt configurations can enter through the same calibration procedure to complement vector reachability. All vector profiles share one post-trained model rather than requiring separate adapters.

Our contributions are:

\begin{enumerate}
\def\labelenumi{\arabic{enumi}.}
\tightlist
\item
  \textbf{A task of constructing and retrieving heterogeneous configurations by multidimensional target scores.} Reachability and coverage require one configuration to satisfy all dimension-wise error requirements.
\item
  \textbf{Evidence on data, refinement, and injection scope in single-axis control.} Matched-data, fixed-layer, norm-matched, and cognitive/behavioral supervision comparisons provide measured components for joint control.
\item
  \textbf{Efficient semantic grids and quantified density/range effects.} Token routing, continuous batching, and safe cache reuse support parallel calibration; density comparisons measure precision gains, and complementary prompts expand reachable regions.
\end{enumerate}

\section*{2 Related Work}\label{related-work}

\subsection*{2.1 Heterogeneous User Simulation and Participant Perspective}\label{heterogeneous-user-simulation-and-participant-perspective}

USP identifies role confusion caused by helpful-assistant training and generates user utterances through implicit profiles, conditional fine-tuning, and cycle-consistency reinforcement learning \citep{ref50}. This motivates learning shared participant-style expression before freezing the model and constructing differentiated configurations.

Sources of heterogeneity include demographics \citep{ref2}, Anthology's backstories \citep{ref11}, Generative Agents' memory and planning \citep{ref3}, and individual records \citep{ref12}. RoleLLM studies role capabilities \citep{ref13}; ThinkingUS models situated user reasoning \citep{ref4}; HER distinguishes first-person role reasoning from third-person model reasoning \citep{ref5}; the Assistant Axis analyzes default assistant identity \citep{ref6}. FPSRG separates the shared participant standpoint from variable attributes, using directions and measured responses to create retrievable profiles.

\subsection*{2.2 Activation Interventions and Multi-Attribute Composition}\label{activation-interventions-and-multi-attribute-composition}

PPLM modifies hidden states with attribute gradients; FUDGE adjusts probabilities using partial-sequence attribute predictions \citep{ref14,ref15}. ITI shifts attention-head activations \citep{ref46}, CAA extracts contrastive residual directions \citep{ref16}, BiPO learns directions through bidirectional preference optimization \citep{ref47}, and ReFT learns parameterized representation interventions \citep{ref17}. We connect contrastive initialization to joint refinement on self-report and behavioral targets.

Composition approaches include dynamic attribute adjustment \citep{ref18}, MAT-Steer's sparse orthogonal interventions \citep{ref48}, K-Steering's multi-label control \citep{ref19}, and PID/local-dynamics feedback \citep{ref20,ref21}. SAS sequentially fits probes under preceding interventions with layer, strength, and quality selection \citep{ref7}; PERSONA uses vector algebra and Persona-Flow \citep{ref8}; ORBIT combines subspace rotations and tokenwise gates \citep{ref9}; Pre-Control tracks attribute scores through a value network and target-error gradients \citep{ref42}.

LoRA adapts low-rank parameters \citep{ref10}, while LoraHub combines adapters using new-task examples \citep{ref30}. FPSRG shares a post-trained model, constructs profiles through activation coefficients, records complete joint responses offline, and retrieves by scores online. Both vectors and prompts can serve as configurations with measured coordinates.

\subsection*{2.3 Semantic Responses and Behavioral Measurement}\label{semantic-responses-and-behavioral-measurement}

MAP-Elites bins behavior descriptors and retains elites through mutation and within-bin competition; Quality-Diversity combines regional diversity and quality \citep{ref27,ref28}. FPSRG uses a quality-constrained catalog, currently measuring a fixed input-coefficient grid, with emphasis on LLM direction construction, joint calibration, and computation reuse. Sparse grids reduce high-dimensional approximation cost under conditions such as mixed smoothness \citep{ref29}, suggesting routes to fewer measurements.

Strength, dynamics, and output quality determine usable intervention ranges \citep{ref7,ref22,ref23}. Persona Dosing inverts targets through isotonic regression and interpolation \citep{ref43}; FPSRG calibrates complete multidimensional configurations and compares reachable regions. Generalized additive models and Belief Dynamics provide link-function and log-odds perspectives \citep{ref44,ref45}; we use these to construct saturation-aware single-axis predictions and evaluate the information added by joint measurement.

Personality inventories \citep{ref24}, ActTraitBench's behavioral paradigms \citep{ref49}, and Li et al.'s Chinese situational judgments \citep{ref25} provide complementary observations. We evaluate within-system coverage, complete-generation remeasurement, and external expression separately, grounding interpretability in the correspondence between dimension definitions, internal operations, and measured responses.

\section*{3 The FPSRG Method}\label{the-fpsrg-method}

FPSRG performs shared first-person post-training, direction learning, joint calibration, and target retrieval. Applications specify dimensions, candidate layers, coefficients, and measurement interfaces; Appendices A--B give instance settings.

\subsection*{3.1 Joint Configuration within an Evaluation System}\label{joint-configuration-within-an-evaluation-system}

Let \(M_\theta\) have \(L\) decoder blocks and width \(H\), with \(d\) dimension definitions:

\[D_k=(\sigma_k,e_k^-,e_k^+,\mathcal X_k,\phi_k),\qquad k\in[d]. \tag{1}\]

Here \(\sigma_k\), \(e_k^-/e_k^+\), \(\mathcal X_k\), and \(\phi_k\) denote semantics, low/high endpoints, situations, and scoring rules. Protocol \(\pi\) fixes the model, items, encodings, sampling, and injection. The joint response to \(a\) is

\[F_\pi(a)=\mathbb E_{Z\sim P_\pi(\cdot\mid M_{\theta_{\rm FP}},a)}[\phi_\pi(Z)]\in\mathbb R^d. \tag{2}\]

The catalog records complete score vectors, including each direction's cross-dimensional effects. Training, development, and calibration sets support learning, layer selection, and measurement of \(\widehat F_\pi\); configurations are fixed before remeasurement.

Coefficients \(a\) and scores \(F_\pi(a)\) occupy different coordinate systems. Raising an extraversion coefficient may change both extraversion and agreeableness. Retrieval uses the complete measured response rather than treating a coefficient as its corresponding trait score.

\subsection*{3.2 Shared First-Person Adaptation and Initialization}\label{shared-first-person-adaptation-and-initialization}

Shared post-training teaches the model to judge and answer as the participant. Records \((x,\rho,y)\) contain a situation, participant-perspective reasoning, and final words or actions; mathematical and long-reasoning examples also concern personal decisions. Mean cross-entropy supervises only current-turn reasoning, answers, and termination tokens. We update LoRA \citep{ref10} or all language parameters, then freeze \(M_{\theta_{\rm FP}}\). Profiles share this model, with vectors supplying attribute differences.

High/low examples share facts and differ in endpoint descriptions or responses. Let \(H_\ell(x)\) be the residual sequence, \(\mathcal P/\mathcal P_0\) the direction/reference pooling operators, \(\mathcal R\) the reference-norm statistic, and \(h_\ell^{\mathcal P}=\mathcal P(H_\ell)\):

\[\begin{aligned}
\delta_{k,\ell}&=\frac{1}{n_k|\mathcal M_k|}\sum_{i=1}^{n_k}\sum_{m\in\mathcal M_k}
\big[h_\ell^{\mathcal P}(x_{k,i,m}^+)-h_\ell^{\mathcal P}(x_{k,i,m}^-)\big],\\
s_{k,\ell}&=\mathcal R\!\left(\left\{\|h_\ell^{\mathcal P_0}(x)\|_2:x\in\mathcal D_k^{\rm ref}\right\}\right),\\
v_{k,\ell}^{(0)}&=\rho_s s_{k,\ell}\frac{\delta_{k,\ell}}{\|\delta_{k,\ell}\|_2},\\
v_{k,\ell}&=\rho_s s_{k,\ell}u_{k,\ell},\quad\|u_{k,\ell}\|_2\le1.
\end{aligned} \tag{3}\]

Encoding index \(m\), reference samples \(\mathcal D_k^{\rm ref}\), layer scale \(s_{k,\ell}\), and budget \(\rho_s\) determine the constraint. We optimize \(u\), project it onto the unit ball after updates, and exclude zero contrasts. Pooling excludes padding and is fixed beforehand. Main non-thinking and HEXACO role directions use last-position pooling and median reference norms; six-dimensional thinking uses mean reasoning contrasts and mean reference norms. Refinement retains learned lengths within budget; Appendix I.1 specifies other baselines.

Equation (3) averages \(n_k\) paired examples and encodings \(\mathcal M_k\), determines the semantic direction, then scales it relative to the layer's reference states. Separate direction/reference pooling accommodates different interfaces under the same norm constraint.

\subsection*{3.3 Cognitive--Behavioral Direction Refinement}\label{cognitivebehavioral-direction-refinement}

For non-thinking measurements, the unsteered high-pole probability \(p_{0,m}\) defines the soft target at dose \(\alpha\):

\[q_{\alpha,m}=\sigma\!\left(\operatorname{logit}[\operatorname{clip}(p_{0,m},\varepsilon,1-\varepsilon)]+\kappa\alpha\right). \tag{4}\]

\(\sigma\) is sigmoid, \(\varepsilon\) clips endpoints, and \(\kappa\) specifies a log-odds shift relative to each item's baseline. \(\mathcal L_{\rm AB}\) is binary cross-entropy on high-pole probability; \(\mathcal L_{\rm self}\) weights length-normalized high/low self-report losses by \(q_{\alpha,m}\) and its complement. Opposite doses supervise complete actions in situation \(c\):

\[\mathcal L_{\rm action}=\frac12\big[
\overline{\operatorname{CE}}(y_B^+\mid c,+\beta v)
+\overline{\operatorname{CE}}(y_B^-\mid c,-\beta v)\big]. \tag{5}\]

Neutral preservation compares unsteered and steered full-vocabulary distributions at identical teacher-forced prefixes. The combined objective is

\[\begin{aligned}
\mathcal L_{\rm vec}^{(n)}
&=\lambda_{\rm AB}\mathcal L_{\rm AB}
+\lambda_{\rm self}\mathcal L_{\rm self}
+\lambda_{\rm action}\mathcal L_{\rm action}\\
&\quad+\lambda_N\chi_n\mathcal L_{\rm neutral}
+\lambda_u\|u\|_2^2.
\end{aligned} \tag{6}\]

The \(\lambda\) terms weight losses, \(\chi_n\) schedules neutral preservation at step \(n\), and \(\beta>0\) is behavioral dose. Thinking interfaces directly supervise complete questionnaire/action demonstrations, with \(q\in\{Q,B\}\) and \(\epsilon\in\{-1,+1\}\):

\[\begin{aligned}
\mathcal L_{\rm trace}=\frac14\sum_{q,\epsilon}\big[&\lambda_\rho\overline{\operatorname{CE}}(\rho_q^\epsilon\mid x_q,\epsilon v)\\
&+\lambda_y\overline{\operatorname{CE}}(y_q^\epsilon\mid x_q,\rho_q^\epsilon,\epsilon v)\big].
\end{aligned}\]

Each segment is normalized by valid length; neutral and norm terms are added. Endpoint instructions generate demonstrations but are absent from vector-training inputs. Both interfaces freeze the model and optimize only directions. Development selection uses bidirectional responses or complete-answer endpoint differences and accounts for invalid outputs (Appendix B).

\subsection*{3.4 Full-Sequence Joint Injection}\label{full-sequence-joint-injection}

The controller \(\mathcal V=\{(\ell_k,v_k)\}_{k=1}^d\) sums same-layer directions:

\[\begin{aligned}
\Delta_\ell(a)&=\sum_{k:\ell_k=\ell}a_kv_k,\\
h'_{\ell,j}&=h_{\ell,j}+m_j\Delta_\ell(a).
\end{aligned} \tag{7}\]

\(m_j=1\) at nonpadding positions. Refinement, calibration, and deployment cover inputs and target/generated positions, including reasoning tokens. Subsequent joint measurement records the resulting coupling.

\subsection*{3.5 Joint Measurement, Filtering, and Retrieval}\label{joint-measurement-filtering-and-retrieval}

Finite coefficient sets \(G_k\) define configurations and their response catalog:

\[\begin{aligned}
\mathcal G&=\prod_{k=1}^dG_k,\qquad N_G=\prod_{k=1}^d|G_k|,\\
\mathcal C_\pi&=\{(a,\widehat F_\pi(a),\eta(a)):a\in\mathcal G\}.
\end{aligned} \tag{8}\]

\(\eta\) retains quality and measurement denominators. Scoring weights are fixed before construction. Both modes support complete generation: parse final answers and aggregate repeated scores. Reasoning, multi-token labels, and free-response evaluation use this interface.

In non-thinking mode, when the first position is the answer and every legal label is one token, one forward pass obtains the conditional distribution on \(\mathcal V_{i,m}\):

\[p^L_{i,m,u}(a)=\frac{\exp(z_{i,m,u}(a)/T)}{\sum_{v\in\mathcal V_{i,m}}\exp(z_{i,m,v}(a)/T)}. \tag{9}\]

\(z\) denotes logits, \(T\) the scoring temperature, and \(i,m,u\) items, encodings, and labels. After restoring scoring keys, expectations are averaged over encodings and item groups, replacing repeated sampling with one pass. Both interfaces retain full-input injection. Conditional expectations and generation means form separate catalogs; cross-interface use requires remeasurement (Appendix B.3).

First-token scoring accelerates answer acquisition without changing item counts or scoring rules. Complete generation also depends on decoding and answer validity. Recording these conditions ties reported scores to the actual calling protocol.

Joint perturbations may disrupt output, while legal-set normalization can hide probability elsewhere. The protocol fixes quality conditions \(\mathcal J_\pi\) and admissible set \(\mathcal G_{\rm adm}=\{a\in\mathcal G:\forall r\in\mathcal J_\pi,\ g_{\pi,r}(\eta(a))=1\}\) before matching targets. The non-thinking cognitive--behavioral interface requires \ensuremath{\geq}98\% mean legal mass per dimension/interface and checks numeric mappings, behavioral order, and label bias. Questionnaire-only evaluation uses the first two conditions; generation checks termination and answer format (Appendices B, I).

Candidates for target \(t\) satisfy maximum coordinate error \(\tau_\infty\) and total error \(\tau_1\):

\[\mathcal G_t=\left\{a\in\mathcal G_{\rm adm}:\begin{aligned}
&\|\widehat F_\pi(a)-t\|_\infty\le\tau_\infty,\\
&\|\widehat F_\pi(a)-t\|_1\le\tau_1
\end{aligned}\right\}. \tag{10}\]

If nonempty, return its nearest-L1 configuration; otherwise return the nearest admissible point and an unmet-target status. Coverage is

\[\operatorname{Cov}(\mathcal T;\tau_\infty,\tau_1)=\frac1{|\mathcal T|}\sum_{t\in\mathcal T}\mathbf1[\mathcal G_t\ne\varnothing]. \tag{11}\]

The service returns configurations, calibration scores, errors, and answers. Changed models or measurement protocols require recalibration.

\subsection*{3.6 Configuration Routing and Cache-Isolated Parallel Inference}\label{configuration-routing-and-cache-isolated-parallel-inference}

Each single-GPU replica retains one model and precomputes per-layer increments for node \(p\). Let \(r(j)\) identify the request owning the \(j\)th scheduled valid token:

\[\begin{aligned}
\mathcal B_{\ell,p}&=\sum_{k:\ell_k=\ell}a_{p,k}v_k,\\
h'_{\ell,j}&=h_{\ell,j}+m_j\operatorname{cast}\!\left(\mathcal B_{\ell,p_{r(j)}}\right).
\end{aligned} \tag{12}\]

Same-layer increments are summed in FP32 and cast to hidden-state precision. Routing follows actual token order through prefill, reasoning, and answers. Different personalities and prompts share continuous batches, with new requests immediately replacing completed ones. Paged KV, hybrid sequence-state management, and chunked prefill reduce memory cost and interleave long prompts with decoding.

Cache identity binds the model, directions, protocol, node, prefix, positions, and mask:

\[\kappa_{r,j}=\operatorname{Hash}\!\left(\operatorname{id}(M_{\theta_{\rm FP}},\mathcal V,\pi),\ p_r,\ x_{r,\le j},\ \operatorname{pos}_{\le j},\ m_{\le j}\right). \tag{13}\]

Intervened states are reused only under identical computation conditions. Earlier first-token measurements additionally reuse boundary states grouped by upstream coefficients (Appendix H). Replicas claim disjoint tasks and preserve request records; recovery does not resample committed results. Current execution uses eager mode and an FP32 output head, without counting disabled CUDA Graphs.

Matching prefixes within a configuration may reuse state; different configurations retain separate post-intervention states even for identical text. Mixed batches thus accommodate different nodes, prompts, and lengths while preserving each request's control conditions.

\subsection*{3.7 Complementary Prompt and Vector Catalogs}\label{complementary-prompt-and-vector-catalogs}

Under the same protocol, measure vector \(\mathcal C_V\), prompt \(\mathcal C_P\), and simultaneous-control \(\mathcal C_{VP}\) catalogs. Catalog complementarity returns one configuration from \(\mathcal C_V\cup\mathcal C_P\); within-generation complementarity measures the increment from adding vectors under fixed text. Both rely on the corresponding measured configurations.

\section*{4 Joint Construction and Component Evaluation}\label{joint-construction-and-component-evaluation}

\subsection*{4.1 Non-Thinking Five- and Ten-Dimensional Construction}\label{non-thinking-five--and-ten-dimensional-construction}

Single-axis curves evaluate directional response, joint coverage evaluates reachability, and nearest-match MAE evaluates retrieval error. Calibration and coverage use the same complete calibration items; separate training/development items learn directions and select layers. Section 5 evaluates reuse on external questions.

Official Qwen3.5-27B \citep{ref31} has 64 decoder blocks and width 5,120. Shared LoRA is frozen after one pass over 1,000 participant-perspective examples, using ``You are a person; answer according to the facts of the scenario.'' Each dimension has 30/6/24 training/development/calibration questionnaire items and 16/4/8 behavioral scenarios. Questionnaires follow IPIP and Johnson \citep{ref32,ref33}. Two numeric mappings and A/B orders are equally weighted. Self-report \(Q\) and behavior \(B\) are converted to 0--100, with \(S=(Q+B)/2\). We use \(S\) unless specified; ablations separate components. Sensory sensitivity is Se.

\textbf{Coverage.} Broad tolerance requires one node with coordinate errors \ensuremath{\leq}20 and total error \ensuremath{\leq}\(12d\); strict tolerance requires \textless10 and \textless{}\(6d\). Coverage counts targets with an admissible match; ordinary-L1 success checks the returned node. Each domain uses 1,000 fixed targets. Thinking complementarity uses coordinatewise tolerance; Appendix C.5 supplements bin occupancy.

\textbf{Five dimensions.} O/C/E/A/N select layers 40/32/32/16/32 from eight candidates (zero-based), with increasing single-axis scores at five doses. Coefficients \(\{-2,-1,0,1,2\}\) give 3,125 configurations and 1,000,000 first-token probes, retaining 1,854 nodes after four-part filtering. Broad coverage in 25--75 is 95.3\%/92.2\% before/after filtering; ordinary-L1 success is 76.1\%, versus 92.2\% with feasible-first retrieval. Under the same metric, zero vectors cover 10.3\% and target-independent random singletons average 5.82\% (Appendix C.6).

O/C/E/A/N denote openness, conscientiousness, extraversion, agreeableness, and neuroticism. Single-axis measurements vary one coefficient with others zero; the joint catalog measures complete five-dimensional responses to simultaneous directions.

\textbf{Ten dimensions.} Adding risk taking, future consequences, prosocial allocation, material values, and sensory sensitivity \citep{ref34,ref35,ref36,ref37,ref38} with the same training proportions and \(\{-1,0,1\}\) yields 59,049 configurations, 37,791,360 probes, and 33,942 qualified nodes. All ten single-axis responses increase. Broad coverage before filtering is 41.5\%/69.3\%/95.4\% in 20--80/25--75/30--70, versus 34.7\%/58.6\%/88.2\% after filtering. Five preselected nodes produce 3,200 complete non-thinking answers: 99.56\% are scoreable, with MAE 3.33 against catalog predictions (Appendix E).

\subsection*{4.2 Thinking-Mode Six-Dimensional Individuals and Interaction}\label{thinking-mode-six-dimensional-individuals-and-interaction}

Huihui Qwen3.8-27B-abliterated uses a 64-layer, width-5,120 Qwen3.5 backbone. We train all language parameters for two epochs on 1,500 first-person examples, then freeze them and learn agreeableness, conscientiousness, neuroticism, risk taking, material values, and future consequences (A/C/N/R/M/T), without loading LoRA. Selected layers are 24/24/32/40/24/24. Three levels \(\{-1,0,1\}\) yield 729 configurations. We generate reasoning and answers but score only answers: 24 questionnaire and eight behavioral items per dimension, two repeats per encoding, give 559,872 outputs; 558,095 are format-valid (99.68\%). Requiring 126/128 valid answers in all dimensions retains 663 nodes (Appendices A.4, B.5).

\textbf{Personality responses.} Eight new Chinese scenarios per dimension use low/high-decile nodes chosen to minimize other-dimension differences, plus middle and zero-vector controls. Forty-eight scenarios × four configurations × no emotion/joy/anger × three samples produce 1,728 complete responses without personality or emotion text prompts. DeepSeek blindly rates final answers and reasoning on a 0--4 scale without configuration labels; 1,702 records receive valid ratings. Same-scenario, same-seed pairs are averaged over three repeats, then equally over fully rated scenarios.

Without emotion, A/C/N/R/M/T high-minus-low mean final-answer differences are 0.11/0.17/0.25/0.95/0.96/0.62; reasoning differences are 0.06/0.21/0.25/0.76/0.96/0.62, using 6/8/8/7/8/7 paired scenarios. All six paired means follow catalog directions, describing selected multidimensional configurations whose remaining coordinates are minimized in difference, not exactly matched.

\textbf{Emotion overlay.} With personality fixed, joy/anger vectors increase corresponding final-answer ratings by 0.23/0.19 and reasoning ratings by 0.27/0.16 over no emotion, using 33/37 fully paired scenarios. Both means follow the intended direction, supporting current-state adjustment over individual profiles.

Personality scores select participants; emotion ratings measure their current expression. These are separate measurements. Paired mean changes in this scenario set show configuration differences in answers and generated reasoning without additional personality reminders.

\subsection*{4.3 Single-Axis Prediction versus Joint Calibration}\label{single-axis-prediction-versus-joint-calibration}

We evaluate node-score prediction and target-retrieval errors when joint measurements are omitted.

\subsubsection*{4.3.1 Constructing Single-Axis Predictions}\label{constructing-single-axis-predictions}

A five-dimensional, five-level grid contains 3,125 configurations but only \(1+5\times4=21\) anchors: zero and four nonzero doses per direction. Each anchor records all five scores, retaining cross-dimensional main effects. For unit coordinate \(e_k\),

\[\widehat F_{\mathrm{add}}(a)=F(0)+\sum_{k=1}^{d}\left[F(a_ke_k)-F(0)\right].\]

Measured single-axis values are added across directions. We compare direct addition, clipping to 0--100, and addition in logit space followed by sigmoid \citep{ref44,ref45}, without reading joint nodes. Prompts use all-middle descriptions as the origin and vary one dimension at a time.

\subsubsection*{4.3.2 Information Added by Joint Measurement}\label{information-added-by-joint-measurement}

Prediction MAE evaluates node scores; target MAE evaluates retrieved scores against requests, averaging over nodes/targets and dimensions (Appendix C.2). Non-thinking vectors and prompts use 1,836 and 3,104 qualified joint nodes. Direct/clipped/logit prediction MAEs are 6.325/6.246/5.244 and 5.304/5.068/3.973. On the same 1,000 targets in 20--80, ordinary-L1 target MAE changes from 9.617 to 7.076 for vector logit prediction versus measured lookup, and from 8.899 to 7.346 for prompts. Joint measurements provide more accurate retrieval coordinates.

A separate four-dimensional thinking catalog holds out each of three generation repeats, calibrating on two and remeasuring on the third (Appendix C.3).

\subsubsection*{4.3.3 Implications for Catalog Construction}\label{implications-for-catalog-construction}

Joint calibration makes combination responses omitted by single-axis predictions available for retrieval. New evaluation systems reuse this workflow by specifying controls and scoring, without first fitting cross-dimensional relationships.

\subsection*{4.4 Complementary Reachable Regions of Prompts and Vectors}\label{complementary-reachable-regions-of-prompts-and-vectors}

\textbf{Catalog complementarity.} The same non-thinking five-dimensional interface measures five-level vectors (−2 to 2), seven-level vectors (−3 to 3), and extremely low/low/middle/high/extremely high prompts. Table 1 uses the same 1,000 targets and four-part filtering; catalog union returns one configuration per target.

\begin{table*}[t]
\caption*{\textbf{Table 1. Complementary coverage (\%).} Broad tolerance is coordinatewise \ensuremath{\leq}20 and total \ensuremath{\leq}60; strict tolerance is \textless10 and \textless30.}
\centering
{\small
\begin{tabular}{@{}
  >{\raggedright\arraybackslash}p{(\linewidth - 10\tabcolsep) * \real{0.1304}}
  >{\raggedleft\arraybackslash}p{(\linewidth - 10\tabcolsep) * \real{0.1739}}
  >{\raggedleft\arraybackslash}p{(\linewidth - 10\tabcolsep) * \real{0.1739}}
  >{\raggedleft\arraybackslash}p{(\linewidth - 10\tabcolsep) * \real{0.1739}}
  >{\raggedleft\arraybackslash}p{(\linewidth - 10\tabcolsep) * \real{0.1739}}
  >{\raggedleft\arraybackslash}p{(\linewidth - 10\tabcolsep) * \real{0.1739}}@{}}
\toprule\noalign{}
\begin{minipage}[b]{\linewidth}\raggedright
Candidate catalog
\end{minipage} & \begin{minipage}[b]{\linewidth}\raggedleft
Configurations
\end{minipage} & \begin{minipage}[b]{\linewidth}\raggedleft
20--80 broad
\end{minipage} & \begin{minipage}[b]{\linewidth}\raggedleft
25--75 broad
\end{minipage} & \begin{minipage}[b]{\linewidth}\raggedleft
20--80 strict
\end{minipage} & \begin{minipage}[b]{\linewidth}\raggedleft
25--75 strict
\end{minipage} \\
\midrule\noalign{}

Vector-5 & 3,125 & 82.7\% & 92.2\% & 32.1\% & 44.0\% \\
Vector-7 & 16,807 & 89.7\% & 96.5\% & 41.0\% & 55.2\% \\
Prompt-5 & 3,125 & 92.0\% & 88.7\% & 13.0\% & 7.4\% \\
Vector-5 \(\cup\) Prompt-5 & 6,250 & \textbf{97.0\%} & \textbf{98.7\%} & 42.3\% & 48.9\% \\
Vector-7 \(\cup\) Prompt-5 & 19,932 & \textbf{98.7\%} & \textbf{99.7\%} & \textbf{49.9\%} & \textbf{59.2\%} \\
\bottomrule
\end{tabular}}
\end{table*}

Five-level vectors plus prompts use 6,250 configurations yet exceed the broad coverage of 16,807 seven-level vectors. In 25--75, strict vector coverage is 44.0\%, prompt coverage 7.4\%, and union 48.9\%. Four-dimensional thinking catalogs cover 9.16\%/15.29\%/22.03\% with vectors/prompts/union on 2,401 targets in 20--80 at coordinate tolerance \ensuremath{\leq}10 (Appendix C.4).

\textbf{Within-generation complementarity.} Adding signed vectors to fixed four-dimensional thinking endpoint prompts moves all eight endpoints further in the intended direction; all four conditions exceed 97\% validity. Appendix C.7 preserves complete scores from 384 responses per condition. Catalog union extends reachability; simultaneous addition tests joint control under fixed text.

\subsection*{4.5 Effects of Data Organization, Refinement, and Injection Scope}\label{effects-of-data-organization-refinement-and-injection-scope}

\textbf{Data and refinement.} The gradient-free comparison uses identical questionnaire/behavior targets, averages high--low content activations, combines groups equally, and matches refined norms. With layers, injection, and items fixed, mean −2-to+2 composite span increases from 30.75 to 40.45, improving all five dimensions; agreeableness questionnaire span changes from −10.82 to +16.79. Direct extraction has larger A/N behavioral spans (Appendix D.2).

\textbf{Cognitive and behavioral supervision.} Holding initialization, layers, 240 updates, and full-sequence injection fixed, we vary only supervision. Joint training achieves the highest composite coverage while preserving both interfaces' reachability (Table 2).

\begin{table*}[t]
\caption*{\textbf{Table 2. Fixed-layer supervision coverage (\%).} The same 10,000 targets in \([25,75]^5\) use filtering and coordinatewise \ensuremath{\leq}20/total \ensuremath{\leq}60 tolerances.}
\centering
{\small
\begin{tabular}{@{}lrrr@{}}
\toprule\noalign{}
Refinement target & Q coverage & B coverage & S coverage \\
\midrule\noalign{}

Q only & 78.76 & 65.75 & 85.64 \\
B only & 58.87 & 84.77 & 82.32 \\
Joint QB & 77.32 & 80.81 & \textbf{91.85} \\
\bottomrule
\end{tabular}}
\end{table*}

\textbf{Injection scope.} With common data and layer-selection rules, mean questionnaire span is 7.88 for response-side injection and 27.16 for input-plus-response injection, larger on all five axes. Both are trained and layer-selected separately, comparing complete strategies (Appendix D.4).

\subsection*{4.6 Grid Density and Coverage Gains within a Fixed Range}\label{grid-density-and-coverage-gains-within-a-fixed-range}

Fixing the model, five directions, full-sequence injection, and 320-probe interface, we vary only density: three levels \(\{-2,0,2\}\), five \(\{-2,-1,0,1,2\}\), and nine at 0.5 intervals from −2 to 2. The nested sets contain 243/3,125/59,049 nodes. Old measurements are reused, new nodes added, and identical 1,000 targets and filtering applied.

\begin{table*}[t]
\caption*{\textbf{Table 3. Density, coverage, and matching error.} Five-dimensional tolerances follow Table 1; MAE is per-axis error of ordinary-L1 retrieval.}
\centering
{\small
\begin{tabular}{@{}
  >{\raggedright\arraybackslash}p{(\linewidth - 10\tabcolsep) * \real{0.1304}}
  >{\raggedleft\arraybackslash}p{(\linewidth - 10\tabcolsep) * \real{0.1739}}
  >{\raggedleft\arraybackslash}p{(\linewidth - 10\tabcolsep) * \real{0.1739}}
  >{\raggedleft\arraybackslash}p{(\linewidth - 10\tabcolsep) * \real{0.1739}}
  >{\raggedleft\arraybackslash}p{(\linewidth - 10\tabcolsep) * \real{0.1739}}
  >{\raggedleft\arraybackslash}p{(\linewidth - 10\tabcolsep) * \real{0.1739}}@{}}
\toprule\noalign{}
\begin{minipage}[b]{\linewidth}\raggedright
Target domain
\end{minipage} & \begin{minipage}[b]{\linewidth}\raggedleft
Levels/axis
\end{minipage} & \begin{minipage}[b]{\linewidth}\raggedleft
Qualified/total nodes
\end{minipage} & \begin{minipage}[b]{\linewidth}\raggedleft
Broad coverage
\end{minipage} & \begin{minipage}[b]{\linewidth}\raggedleft
Strict coverage
\end{minipage} & \begin{minipage}[b]{\linewidth}\raggedleft
Per-axis MAE
\end{minipage} \\
\midrule\noalign{}

25---75 & 3 & 120/243 & 71.2\% & 15.6\% & 8.01 \\
25---75 & 5 & 1,854/3,125 & 92.2\% & 44.0\% & 5.78 \\
25---75 & 9 & 38,932/59,049 & \textbf{94.6\%} & \textbf{54.6\%} & \textbf{4.91} \\
20---80 & 3 & 120/243 & 59.1\% & 10.5\% & 9.14 \\
20---80 & 5 & 1,854/3,125 & 82.7\% & 32.1\% & 7.08 \\
20---80 & 9 & 38,932/59,049 & \textbf{87.2\%} & \textbf{40.1\%} & \textbf{6.28} \\
\bottomrule
\end{tabular}}
\end{table*}

In 25--75, three to five levels adds 21.0 percentage points of broad coverage; five to nine adds 2.4. The latter raises strict coverage by 10.6 points and reduces MAE by 15.1\%, using 18.90 times as many nodes. Density improves fine matching; range expansion and prompts add response regions, without retraining directions.

Nine levels reuse 17,308 nodes and add 41,741 with 16 replicas in about ten hours including aggregation. Paired benchmarks retaining identical batches, precision, and operator optimizations show warm cross-node-reuse speedups of 2.76/2.77 and fill-inclusive speedup 1.74, with identical probabilities and legal mass (Appendix H).

\section*{5 External Scenario Transfer and Control-Method Comparisons}\label{external-scenario-transfer-and-control-method-comparisons}

We test external configuration reuse and compare controls on a common backbone and budget.

\subsection*{5.1 Scenario Transfer on Chinese PersonalityBench}\label{scenario-transfer-on-chinese-personalitybench}

We translate the 459 PersonalityBench scenarios used by PERSONA \citep{ref8} into Chinese, preserving identities and meaning: 90/89/90/90/100 items for O/C/E/A/N. Section 4's model, LoRA, vectors, and full-sequence injection are unchanged. Among 1,854 jointly qualified nodes, each dimension selects minimum/maximum \(S\) and the node nearest their midpoint, without constraining other dimensions or reselecting after generation.

Thinking is disabled and no personality prompts are added, producing 1,377 responses. DeepSeek sees only questions and responses, blindly rating 1--5 trait intensity without configurations or calibration scores. This independent scenario scale is not ranked against the original leaderboard. Interpretable evidence within 512 tokens is scored; 32 records lack trait evidence and are excluded, with all others retained. Table 4 reports denominators (Appendix F.3).

\begin{table*}[t]
\caption*{\textbf{Table 4. Chinese PersonalityBench transfer.} Calibration triples denote low/middle/high; parentheses give valid rating counts.}
\centering
{\small
\begin{tabular}{@{}
  >{\raggedright\arraybackslash}p{(\linewidth - 8\tabcolsep) * \real{0.1667}}
  >{\raggedright\arraybackslash}p{(\linewidth - 8\tabcolsep) * \real{0.1667}}
  >{\raggedleft\arraybackslash}p{(\linewidth - 8\tabcolsep) * \real{0.2222}}
  >{\raggedleft\arraybackslash}p{(\linewidth - 8\tabcolsep) * \real{0.2222}}
  >{\raggedleft\arraybackslash}p{(\linewidth - 8\tabcolsep) * \real{0.2222}}@{}}
\toprule\noalign{}
\begin{minipage}[b]{\linewidth}\raggedright
Dimension
\end{minipage} & \begin{minipage}[b]{\linewidth}\raggedright
Calibration scores (0--100)
\end{minipage} & \begin{minipage}[b]{\linewidth}\raggedleft
Low scenario rating
\end{minipage} & \begin{minipage}[b]{\linewidth}\raggedleft
Middle scenario rating
\end{minipage} & \begin{minipage}[b]{\linewidth}\raggedleft
High scenario rating
\end{minipage} \\
\midrule\noalign{}

Openness O & 11.33/46.92/82.58 & 1.63 (87) & 2.72 (85) & 3.81 (80) \\
Conscien\-tious\-ness C & 11.65/47.78/83.96 & 1.71 (87) & 3.33 (85) & 4.22 (88) \\
Extraversion E & 8.22/46.26/84.30 & 1.21 (90) & 2.60 (90) & 4.05 (87) \\
Agree\-able\-ness A & 32.04/57.55/83.08 & 1.96 (90) & 4.28 (90) & 4.12 (89) \\
Neuroticism N & 16.32/46.78/77.18 & 3.38 (98) & 4.23 (99) & 4.95 (100) \\
\bottomrule
\end{tabular}}
\end{table*}

High exceeds low on all five dimensions. O/C/E/N increase across three levels; A's middle slightly exceeds high. Same-item, scoreable high--low pairs average 2.14/2.52/2.85/2.16/1.57, showing same-direction expression of catalog differences in external free responses.

\subsection*{5.2 Single-Axis Control Methods on a Shared Backbone}\label{single-axis-control-methods-on-a-shared-backbone}

Public IPIP HEXACO covers H/E/X/A/C/O, with 240 items split by facet into 144 training, 48 development, and 48 held-out items. CAA \citep{ref16}, an independent probe \citep{ref7}, role initialization, adapted PERSONA \citep{ref8}, BiPO \citep{ref47}, and refinement share Qwen3.5-27B, first-person LoRA, candidate layers, and full-sequence injection. Role initialization uses last-position contrasts and median reference norms. Direct vectors have length 0.1 times the reference scale; BiPO/refinement retain learned lengths within budget. Methods select layers independently on development items; held-out items do not enter optimization. Refinement here uses questionnaires only. The probe corresponds to SAS's single-axis component; other adaptations appear in Appendix I.1.

With thinking disabled, eight held-out items × two mappings × three samples give 48 complete answers per dimension/strength. Keyed 1--5 ratings are converted to 0--100 and averaged equally. PERSONA's negative half-axis uses its independent low-pole vector.

\begin{figure*}[t]
\centering
\pandocbounded{\includegraphics[keepaspectratio,alt={Figure 1. Complete-answer curves for six methods over a fixed interval. Means use parseable answers; refined X/C/O have missing responses at some strong endpoints. Appendix I.2--I.3 reports extensions and repeats.}]{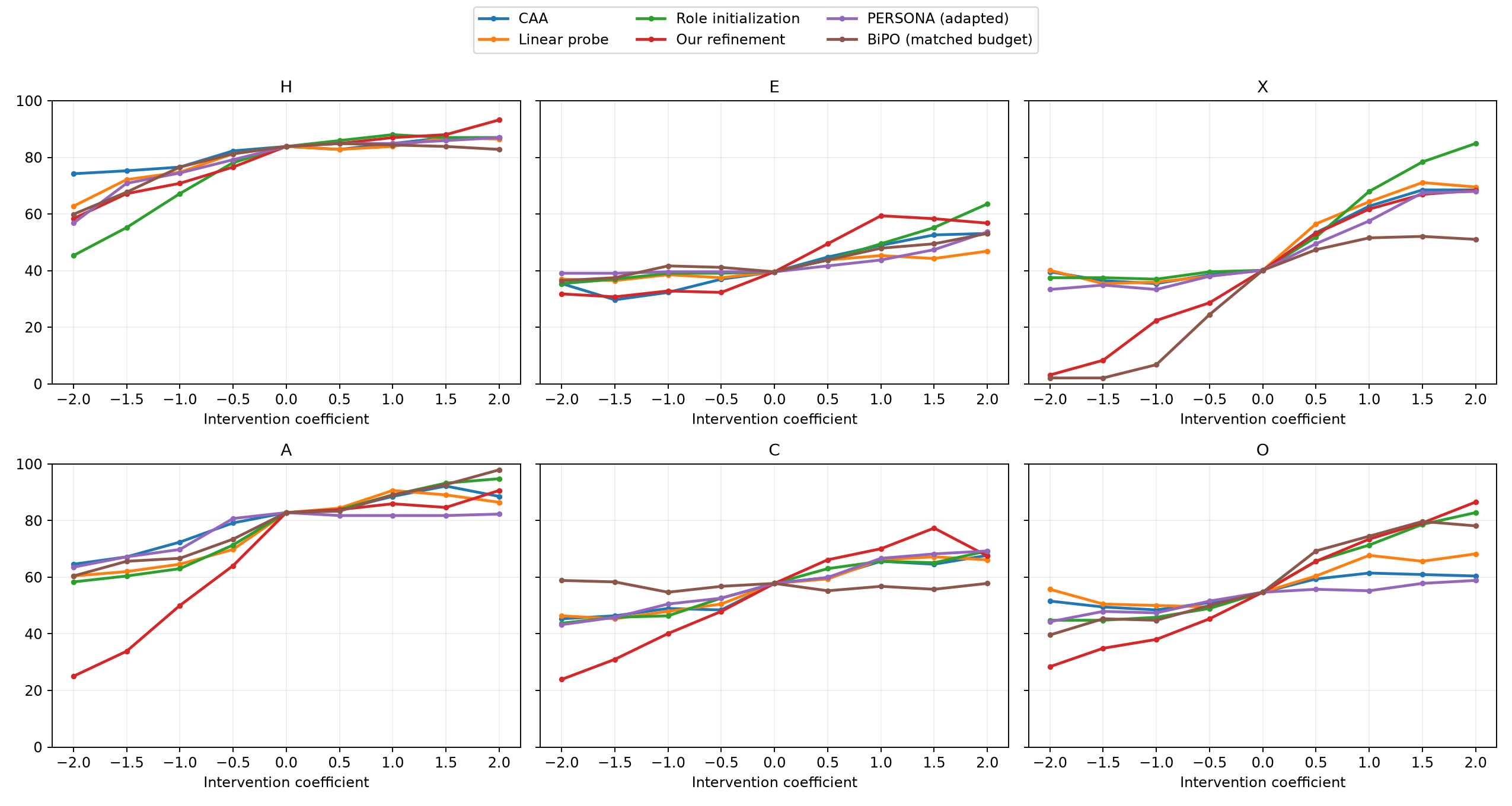}}
\caption*{Figure 1. Complete-answer curves for six methods over a fixed interval. Means use parseable answers; refined X/C/O have missing responses at some strong endpoints. Appendix I.2--I.3 reports extensions and repeats.}
\end{figure*}

Mean −2-to+2 endpoint differences are 48.75 for refinement, 36.20 initialization, 27.26 BiPO, 23.13 PERSONA, 20.23 probe, and 19.10 CAA. Refinement leads on X/A/C/O and initialization on H/E. Agree\-able\-ness spans 25.00--90.63 after refinement versus 58.33--94.79 initially, primarily extending downward control.

Extending both directions until complete-answer validity falls below 90\%, the observed valid-point ranges before first failure average 50.95/45.14/41.23/37.72/29.99/29.47 in that method order. Additional boundary checks leave refinement highest at 50.78 (Appendix I.2--I.3).

\subsection*{5.3 Joint Coverage at a Matched Measurement Budget}\label{joint-coverage-at-a-matched-measurement-budget}

With directions/layers fixed, each method measures 729 combinations at −1/0/+1. Forty-eight held-out items × two mappings give 96 first-token conditional-expectation probes per node. Each dimension requires \ensuremath{\geq}98\% mean legal mass and \ensuremath{\leq}25 mapping-score difference. On the same 10,000 targets in \([25,75]^6\), strict coverage requires per-axis error \textless10 and total \textless36; MAE is per-axis error of nearest-L1 retrieval.

\begin{table*}[t]
\caption*{\textbf{Table 5. Six-dimensional joint results at 729 nodes per method.} Scores follow quality filtering.}
\centering
{\small
\begin{tabular}{@{}
  >{\raggedright\arraybackslash}p{(\linewidth - 6\tabcolsep) * \real{0.2000}}
  >{\raggedleft\arraybackslash}p{(\linewidth - 6\tabcolsep) * \real{0.2667}}
  >{\raggedleft\arraybackslash}p{(\linewidth - 6\tabcolsep) * \real{0.2667}}
  >{\raggedleft\arraybackslash}p{(\linewidth - 6\tabcolsep) * \real{0.2667}}@{}}
\toprule\noalign{}
\begin{minipage}[b]{\linewidth}\raggedright
Construction
\end{minipage} & \begin{minipage}[b]{\linewidth}\raggedleft
Qualified nodes
\end{minipage} & \begin{minipage}[b]{\linewidth}\raggedleft
Strict coverage (\%)↑
\end{minipage} & \begin{minipage}[b]{\linewidth}\raggedleft
Matching MAE↓
\end{minipage} \\
\midrule\noalign{}

CAA & 561 & 4.94 & 10.20 \\
Independent probe & 393 & 4.23 & 10.81 \\
Role initialization & 709 & 9.08 & 8.70 \\
Adapted PERSONA & 728 & 5.31 & 10.04 \\
BiPO & 617 & 3.95 & 10.39 \\
\textbf{Refined (ours)} & 570 & \textbf{12.56} & \textbf{8.13} \\
\bottomrule
\end{tabular}}
\end{table*}

Refinement improves strict coverage by 38.3\% and reduces MAE by 6.6\% relative to initialization. Unfiltered coverage also leads (13.48\% versus 9.37\%). It attains the highest strict coverage and lowest error in 20--80 and 30--70 (Appendix I.4), improving both single-axis span and joint matching under the common protocol.

\subsection*{5.4 Interactive Application with Heterogeneous Market Participants}\label{interactive-application-with-heterogeneous-market-participants}

We connect the thinking-mode four-dimensional catalog to a multi-turn bargaining environment, turning statically calibrated configurations into market participants that observe, deliberate, and act. Four vector-only profiles are selected by relatively low/high measured risk taking and material values, without personality prompts or separate model training for each individual. Frozen 27B participants negotiate price, delivery, and warranty with original or fine-tuned 9B sellers in the same ten trading worlds, producing 80 episodes. Configurations share product facts, budgets, and utility rules. Eight worlds have one buyer and one seller; two have two buyers and two sellers, with the other frozen participants also using the assigned configuration.

\textbf{Heterogeneity appears in the trade-offs made during negotiation.} Participants can maintain an offer, seek better terms, accept a certain return, or leave; generated reasoning and structured actions record these choices together. In the office-headset world with the fine-tuned seller, two higher-material-value profiles face identical valid offers: the seller requests CNY 100 with seven-day delivery and no warranty, while the buyer's outstanding offer is CNY 100 with three-day delivery and a twelve-month warranty. The lower-risk profile waits, explaining, ``If s0 rejects or makes a new offer, I will decide then'' (translated from Chinese). The higher-risk profile retains its delivery and warranty requirements but lowers its offer to CNY 90, probing for better terms. Both ultimately trade on the existing CNY 100 contract, but take different negotiation paths: one maintains its proposal while the other actively seeks a larger return.

Configurations also express different demands for transaction terms. For the same product, another profile says at the opening that ``a warranty is optional'' (translated) and offers CNY 110 with three-day delivery and no warranty; the higher-material-value profile above retains twelve months of warranty in its subsequent deliberation. These records illustrate local differences in service demands, waiting, and counteroffers among callable individuals. They are interaction cases: service demands are not equated with material values, and no single action is treated as a fixed rule for high or low risk taking. Behavior emerges jointly from the configuration, economic conditions, and dialogue history.

This application connects score-based selection, participant invocation, negotiation, and settlement. Users can choose individuals with different calibrated characteristics from one model's catalog, construct a heterogeneous population in a shared environment, and inspect their reasons, actions, and outcomes. Fine-tuning the 9B seller on interaction data increases mean contractual utility from CNY 13.835 to 20.725 (+49.8\%) across forty evaluations in the same ten test worlds.

\section*{6 Conclusion and Future Work}\label{conclusion-and-future-work}

FPSRG combines shared first-person post-training, direction learning, and joint measurement to construct heterogeneous profiles retrievable by multidimensional scores. Non-thinking five/ten-dimensional and thinking six-dimensional instances implement the workflow. Data, refinement, and injection comparisons support single-axis improvements; the public six-dimensional experiment also shows joint-coverage and matching gains. Joint calibration records combination responses, density improves precision, prompts extend reachability, and token routing with isolated caches supports parallel use.

Chinese PersonalityBench separates high/low profiles in the intended direction on all five axes, with three-level increases on four. Applications can define assessments, build catalogs, and invoke configurations in interactive simulations. Future work addresses fewer measurements and cross-dimensional response mechanisms.

\section*{7 Ethics, Scope, and Reproducibility}\label{ethics-scope-and-reproducibility}

We construct simulated participants; deployments should disclose model identity. Scores are not for real-person diagnosis or nonconsensual deception/manipulation. Catalog coverage, same-item remeasurement, and external scenarios evaluate different interfaces. Interpretability refers to traceable dimensions, operations, and responses; generated reasoning need not faithfully describe hidden computation \citep{ref26}.

Thinking and non-thinking use separate catalogs. Changed models, questions, or scoring require recalibration or independent checks. Reproduction materials contain splits, training/layer selection, model/direction identities, scoring/filtering, node outputs, and sources. Resources follow their licenses; formal submission will include public archives, contributions, funding, and applicable review information.

%% file: appendix.tex
\section*{Appendices}\label{appendices}

The following appendices document data, optimization, scoring, coverage, remeasurement, external evaluation, and execution. Training, development, calibration, and independent repeated generation serve different roles throughout.

\section*{Appendix A Data, Dimensions, and Shared Adaptation}\label{appendix-a-data-dimensions-and-shared-adaptation}

\subsection*{A.1 Data Roles and Dimension Definitions}\label{a.1-data-roles-and-dimension-definitions}

Each dimension contains 60 questionnaire items, split into 30 for direction learning, six for development selection, and 24 for joint calibration. Twenty-eight behavioral scenarios are split 16/4/8, with event-family checks across splits. Ten dimensions therefore comprise 600 questionnaire items and 280 behavioral scenarios. Big Five items follow the IPIP framework; the additional constructs use the references in Section 4.1. Chinese items operationalize these constructs for this study.

The high poles are: O, exploring new experiences and ideas; C, fulfilling responsibilities and checking work; E, initiating social interaction and expression; A, cooperation and consideration with personal boundaries; N, worry and emotional reactivity under stress; R, accepting uncertainty with specified probabilities and affordable losses; T, prioritizing delayed outcomes; V, voluntarily exchanging limited personal benefit for others' or collective benefit; M, relating possessions to goals, happiness, and success; and sensory Se, sensitivity to sound, light, smell, touch, and concurrent stimulation. Se denotes the sensory trait; S denotes only the composite score.

Scores are model responses under the specified tasks, not population-norm diagnoses. Direction learning and calibration use different items. Exposure during shared post-training is recorded separately, including the six-dimensional consumer instance's overlap in Appendix A.4. Catalog scores are distinguished from subsequent independent generations.

\subsection*{A.2 First-Person Adaptation}\label{a.2-first-person-adaptation}

The shared dataset contains 1,000 unique targets in 151 dialogue groups: 280 everyday-life and negotiation examples, 397 mathematics and extended-reasoning examples, and 323 participant decisions. Former tool interactions are rewritten as participant reasoning and conversational responses based on scenario facts. Supervision includes the current reasoning, reasoning terminator, final answer, and end-of-turn token, excluding history and context. There are 349,660 target tokens; the maximum sequence is 4,338 tokens under an 8,192-token training limit, with no target truncation.

\textbf{Table A1. Shared adaptation settings.}

{\small\begin{longtable}[]{@{}
  >{\raggedright\arraybackslash}p{(\linewidth - 2\tabcolsep) * \real{0.5000}}
  >{\raggedright\arraybackslash}p{(\linewidth - 2\tabcolsep) * \real{0.5000}}@{}}
\toprule\noalign{}
\begin{minipage}[b]{\linewidth}\raggedright
Parameter
\end{minipage} & \begin{minipage}[b]{\linewidth}\raggedright
Setting
\end{minipage} \\
\midrule\noalign{}
\endhead
\bottomrule\noalign{}
\endlastfoot
Original model & Official Qwen/Qwen3.5-27B \\
Initialization & Original weights with a new LoRA; no preceding adapter \\
LoRA rank/scale/dropout & 16/32/0 \\
Optimizer & AdamW \\
Learning rate & \(10^{-5}\); 10 warm-up steps, then linear decay to 0.1 of initial rate \\
Weight decay/gradient clipping & 0.01/1 \\
Global batch & 8; eight replicas, microbatch one each \\
Training passes & One full pass, 125 updates \\
Loss & Mean cross-entropy over current target tokens in the global batch \\
Reference KL/segment weights & Neither \\
Main checkpoint & Update 125, frozen thereafter \\
Direction/grid precision & BF16 backbone + unmerged FP32 LoRA \\
\end{longtable}}

All current target tokens have equal weight.

\subsection*{A.3 Symmetric Behavioral Targets}\label{a.3-symmetric-behavioral-targets}

The approved openness example places the participant and a friend in a pottery workshop. Familiar brush painting and unfamiliar sponge stippling have both been demonstrated, cost the same, take ten minutes, and are equally safe and feasible. There is no competition or gift obligation. The shared instruction asks for first-person speech, action, or both. The high target chooses the new sponge technique to explore its texture; the low target chooses the familiar brush technique and completes the same plate. Scenario resources and task demands are unchanged.

The following Chinese targets are retained verbatim, rather than represented as original English generations:

\begin{quote}
High: {\cjkfont 我把海绵移到碟子旁，先蘸一点颜料试出疏密，再沿碟边点出一圈。}``{\cjkfont 今天我用这个新办法。刷子我已经很熟了，这次想亲手试试点色，看海绵落下去会留下怎样的纹理，再把整只碟子做完。}''

Low: {\cjkfont 我把刷子移到碟子旁，先蘸一点颜料试好厚薄，再沿碟边均匀刷开。}``{\cjkfont 今天我还是用熟悉的平涂。海绵工具先放回去，这次就按我习惯的手法，把整只碟子的颜色铺得匀一些。}''
\end{quote}

English translations of the Chinese targets (for readers, not English model outputs):

\textbf{High.} I move the sponge beside the plate, dip it lightly in paint to try different densities, and stipple a ring around the rim. ``Today I will use this new technique. I already know the brush well; this time I want to try stippling myself, see what texture the sponge leaves, and then finish the whole plate.''

\textbf{Low.} I move the brush beside the plate, dip it lightly in paint to check the thickness, and spread it evenly around the rim. ``Today I will stick to the familiar flat painting. I will put the sponge back and use my usual technique to cover the whole plate more evenly.''

Questionnaire extraction uses high/low role prefixes and both A/B orders, producing 120 role-conditioned and 60 reference inputs per dimension. Catalog measurement removes those prefixes, using two numeric mappings and both behavioral orders. Appendix H.4 indexes the exact templates and item identities.

\subsection*{A.4 Full-Parameter Post-Training for the Six-Dimensional Consumer Instance}\label{a.4-full-parameter-post-training-for-the-six-dimensional-consumer-instance}

The Huihui instance starts from its original 27B weights and uses the original 1,000 participant examples, 250 native-reasoning examples with explicit participant identity, and 250 Chinese conflict-and-boundary examples. The identity examples are different outputs from 96 ACNR questionnaire items. The conflict-and-boundary set comprises 150 situations involving offense, 50 boundary-setting examples, and 50 ordinary exchanges or de-escalation after an apology. These examples train fact-grounded participant responses, not aggression in every situation.

All language-model parameters are updated for two epochs: 188 steps and 3,000 actual example exposures. Unused vision parameters are not updated. The global batch size is 16; AdamW uses learning rate \(5\times10^{-6}\), ten warm-up steps, weight decay 0.01, and gradient clipping at one. Training uses FSDP sharding, FP32 master parameters and optimizer states, BF16 forward computation, activation checkpointing, and chunked vocabulary loss. Supervision covers complete reasoning, final answers, and termination tokens, with no target truncation or example packing. Training takes 54 min 28 s including loading and complete weight exports after both epochs. The second-epoch model is then frozen for new direction learning and catalog construction.

The 96 ACNR source items also appear in six-dimensional calibration; no exact source-item overlap is found for M/T. This instance measures static responses within a specified evaluation system, rather than claiming a test entirely held out from shared post-training. Appendix H's six-dimensional evidence record provides model, data, and completion digests.

\section*{Appendix B Refinement, Selection, and Measurement}\label{appendix-b-refinement-selection-and-measurement}

\subsection*{B.1 Complete Losses and Optimization}\label{b.1-complete-losses-and-optimization}

The choice and self-report losses are

\[\mathcal L_{\rm AB}=\mathbb E_m[-q_{\alpha,m}\log p_{\alpha,m}-(1-q_{\alpha,m})\log(1-p_{\alpha,m})],\]

\[\mathcal L_{\rm self}=\mathbb E_m[q_{\alpha,m}\overline{\operatorname{CE}}(y^+_{Q,m}\mid x_m,\alpha v)+(1-q_{\alpha,m})\overline{\operatorname{CE}}(y^-_{Q,m}\mid x_m,\alpha v)].\]

Length normalization and neutral preservation are

\[\overline{\operatorname{CE}}(y\mid x,\alpha v)=-\frac1{|y|}\sum_j\log p_{\theta_{\rm FP},\alpha v}(y_j\mid x,y_{<j}),\]

\[\mathcal L_{\rm neutral}=\frac1{|\mathcal J_N|}\sum_{j\in\mathcal J_N}D_{\rm KL}\big(p_{\theta_{\rm FP},0}(\cdot\mid x_N,y_{N,<j})\,\|\,p_{\theta_{\rm FP},\alpha_Nv}(\cdot\mid x_N,y_{N,<j})\big).\]

Each questionnaire encoding has its own baseline and equal weight. Complete targets include termination tokens and are normalized by their own valid lengths. Candidate zero-based layers are 8/16/24/32/40/48/56/60. The relative norm budget is 0.1. Adam uses learning rate 0.003 for 240 updates with gradient clipping at one. Questionnaire doses cycle through −1, −0.5, 0.5, 1; scoring temperature is 0.7 and log-odds displacement coefficient is two. Behavioral strength alternates between 0.5 and one after each 16-scenario cycle, with opposite signs for high/low targets.

Loss weights for choice, self-report, behavior, neutral preservation, and norm penalty are 0.5/0.1/0.5/0.1/0.01. Neutral preservation runs every four updates with alternating signed neutral doses. Each step includes both questionnaire orders and one behavioral pair. Projection is \(u\leftarrow u/\max(1,\|u\|_2)\). Fixed cycling periods are 30 questionnaire items and 16 behavioral scenarios. Questionnaire-only removes the action loss; behavior-only removes choice and self-report losses without rescaling remaining weights.

\subsection*{B.2 Development Selection}\label{b.2-development-selection}

Development doses are −1, −0.5, 0, 0.5, and 1. The questionnaire signal is the restored expected numeric score; the behavioral signal is the difference between length-normalized high/low target log likelihoods. For interface \(m\in\{Q,B\}\), the weaker bidirectional effect is

\[b^m_{k,\ell}=\min\{s^m_{k,\ell}(a_{\rm dev})-s^m_{k,\ell}(0),\ s^m_{k,\ell}(0)-s^m_{k,\ell}(-a_{\rm dev})\}.\]

Convert effects to midrank percentiles across candidate layers. Selection maximizes, in order, the smaller interface percentile, mean percentile, behavioral adjacent monotonicity, and negative added questionnaire mapping discrepancy, then favors the shallower layer. With \(c\) candidates, midrank is \((\#\{b'<b\}+0.5(\#\{b'=b\}-1))/(c-1)\). Adjacent positive/zero/negative changes receive 1/0.5/0. A single candidate is adopted directly.

\textbf{Table B1. Selected layers for the non-thinking main catalogs.}

{\small\begin{longtable}[]{@{}lrrrrrrrrrr@{}}
\toprule\noalign{}
Dimension & O & C & E & A & N & R & T & V & M & Se \\
\midrule\noalign{}
\endhead
\bottomrule\noalign{}
\endlastfoot
Main catalog layer & 40 & 32 & 32 & 16 & 32 & 32 & 40 & 32 & 40 & 32 \\
\end{longtable}}

\subsection*{B.3 Scores and Quality Filtering}\label{b.3-scores-and-quality-filtering}

Questionnaire expectations over legal labels 1--5 are restored to 0/25/50/75/100 using mapping and polarity keys. Behavioral scores are high-pole A/B probability multiplied by 100. Five- and ten-dimensional nodes use 320 and 640 probes at temperature 0.7. S averages Q and B equally. Full-vocabulary legal mass is

\[\mu_{i,m}(a)=\frac{\sum_{u\in\mathcal V_{i,m}}\exp(z_{i,m,u}(a)/T)}{\sum_{u\in\mathcal V}\exp(z_{i,m,u}(a)/T)}.\]

All four conditions must hold before retrieval:

\begin{enumerate}
\def\labelenumi{\arabic{enumi}.}
\tightlist
\item
  Mean legal mass is at least 0.98 for every dimension and each Q/B interface.
\item
  Mean absolute questionnaire score discrepancy between restored mappings is at most 25.
\item
  Mean absolute behavioral high-pole probability discrepancy between orders is at most 0.25.
\item
  The raw A-label mean satisfies \(\max(\bar p_A,1-\bar p_A)\le0.90\).
\end{enumerate}

The first criterion is coordinate-specific; the other three aggregate the node. The 98\% threshold budgets at most 2\% mean non-answer probability. Same-layer inner products contribute to joint perturbation:

\[\|\Delta_\ell(a)\|_2^2=\sum_{k\in I_\ell}a_k^2\|v_k\|_2^2+2\sum_{k<k';\ k,k'\in I_\ell}a_ka_{k'}\langle v_k,v_{k'}\rangle.\]

Thus output quality is measured directly. Calibration-interface filtering and full-dialogue quality are evaluated separately; failed nodes remain in the raw catalog.

\subsection*{B.4 Directions and Construction of the Original Four-Dimensional Thinking Catalog}\label{b.4-directions-and-construction-of-the-original-four-dimensional-thinking-catalog}

The same adapter supplies high/low demonstrations generated by the frozen model. Only complete, target-consistent pairs are retained. Initialization averages hidden states over demonstration reasoning and scales the difference to 0.1 times the mean reference-state norm. Each step contains questionnaire and behavioral high/low examples at coefficients ±1. Reasoning/final-answer length-normalized cross-entropy weights are 0.7/0.3. Adam uses 0.003 for 240 updates, clipping one, and L2 weight 0.01. Signed neutral KL runs every four updates with weight 0.05 per sign.

Development selection uses mean directed Q/B separation minus invalidity, prioritizing 98\% development validity. Eight-dimensional single-axis development generates 9,216 responses. The O/R/M/A catalog uses layers 32/16/24/16, 81 nodes, 128 items, and three repeats, with fixed encoding, medium thinking, temperature 0.7, and 2,048 tokens. Vector and prompt catalogs have answer-validity rates 96.80\% and 99.28\%. Generation validity is distinct from non-thinking legal-probability gating.

\textbf{Table B2. Joint responses by coefficient level in the original four-dimensional thinking catalog.} Holding the focal coefficient fixed, values aggregate the other three dimensions' 27 combinations and three repeats. This catalog supports Sections 4.3--4.4, not the six-dimensional full-parameter model.

{\small\begin{longtable}[]{@{}lrrr@{}}
\toprule\noalign{}
Dimension & Coefficient −1 & Coefficient 0 & Coefficient +1 \\
\midrule\noalign{}
\endhead
\bottomrule\noalign{}
\endlastfoot
Openness O & 54.25 & 62.91 & 69.20 \\
Risk taking R & 29.94 & 32.44 & 36.61 \\
Material values M & 16.83 & 27.15 & 43.34 \\
Agree\-able\-ness A & 59.92 & 65.43 & 72.82 \\
\end{longtable}}

\subsection*{B.5 Refinement and Filtering in the Six-Dimensional Thinking Instance}\label{b.5-refinement-and-filtering-in-the-six-dimensional-thinking-instance}

The frozen, fully post-trained model generates 1,104 high/low training demonstrations and eight neutral anchors. Training demonstrations alone allow at most two logged format-repair attempts. A total of 1,549 attempts yields 548 complete high/low pairs; invalid formal development or grid outputs are not resampled. For each axis, the first four complete questionnaire pairs and first four behavioral pairs initialize the direction. Mean reasoning activations are extracted under teacher-forced inputs without high/low role instructions. Their high--low difference is scaled to 0.1 times the mean norm of the corresponding reasoning-mean vectors.

Each of the six dimensions is refined at zero-based layers 8/16/24/32/40/48/56/60 for 240 steps with learning rate 0.003. Each step contains one high and one low example from both questionnaire and behavior, with coefficients ±1. Reasoning and final-answer losses are normalized separately by length and weighted 0.7/0.3. Bilateral neutral KL is applied every four steps with total weight 0.1, and the vector L2 weight is 0.01. Parameters are projected onto the unit ball after each update. Injection covers the full input and target sequence.

Development evaluation generates 2,880 answers at −1/0/+1. It first prioritizes candidates with validity at least 98\%, then ranks the equal-weighted questionnaire/behavior endpoint difference minus \(100(1-\text{validity})\), breaking ties toward shallower layers. Invalid development outputs receive a temporary score of 50 and the separate penalty; formal scores do not impute invalid answers. All six selected directions have 100\% development validity. This is layer-selection measurement, not an additional complete single-axis scan.

The formal catalog uses 24 questionnaire and eight behavioral items per dimension, with two repeats of each encoding, medium thinking, temperature 0.7, and a 4,096-output-token cap. Within each repeat, encodings and Q/B components receive equal weight; scores are then averaged over the two repeats. Invalid outputs remain in each dimension's denominator of 128 answers. Requiring at least 126 valid answers in every dimension retains 663 nodes. Deployment separately requires a literal occurrence rate of the Chinese word ``{\cjkfont 用户}'' (user) no greater than 10\%---all nodes pass in this run---and removes six qualified nodes on an existing exclusion list, leaving 657 serving candidates. This deployment exclusion changes neither the original 729 scores nor the 663 format-qualified count. The lexical condition is not a semantic certification of first-person reasoning.

\section*{Appendix C Coverage, Retrieval, and Joint Calibration}\label{appendix-c-coverage-retrieval-and-joint-calibration}

\subsection*{C.1 Targets and Full Coverage}\label{c.1-targets-and-full-coverage}

Primary coverage uses 1,000 ten-dimensional uniform unit draws with seed 20261004; five-dimensional analyses use the first five columns and rescale to each target domain. Feasibility requires per-axis error at most 20 and total at most \(12d\) at the same node. Existence coverage and ordinary-L1 returned-node success are distinct. Refinement ablations use 10,000 targets per domain.

\textbf{Table C1. Composite dimension-score coverage and returned-node success across target domains (\%).}

{\small\begin{longtable}[]{@{}
  >{\raggedright\arraybackslash}p{(\linewidth - 10\tabcolsep) * \real{0.1364}}
  >{\raggedright\arraybackslash}p{(\linewidth - 10\tabcolsep) * \real{0.1364}}
  >{\raggedleft\arraybackslash}p{(\linewidth - 10\tabcolsep) * \real{0.1818}}
  >{\raggedleft\arraybackslash}p{(\linewidth - 10\tabcolsep) * \real{0.1818}}
  >{\raggedleft\arraybackslash}p{(\linewidth - 10\tabcolsep) * \real{0.1818}}
  >{\raggedleft\arraybackslash}p{(\linewidth - 10\tabcolsep) * \real{0.1818}}@{}}
\toprule\noalign{}
\begin{minipage}[b]{\linewidth}\raggedright
Catalog
\end{minipage} & \begin{minipage}[b]{\linewidth}\raggedright
Domain
\end{minipage} & \begin{minipage}[b]{\linewidth}\raggedleft
Unfiltered coverage
\end{minipage} & \begin{minipage}[b]{\linewidth}\raggedleft
Qualified coverage
\end{minipage} & \begin{minipage}[b]{\linewidth}\raggedleft
Unfiltered L1 success
\end{minipage} & \begin{minipage}[b]{\linewidth}\raggedleft
Qualified L1 success
\end{minipage} \\
\midrule\noalign{}
\endhead
\bottomrule\noalign{}
\endlastfoot
5D & 20---80 & 87.6 & 82.7 & 67.4 & 61.9 \\
5D & 25---75 & 95.3 & 92.2 & 82.0 & 76.1 \\
5D & 30---70 & 99.6 & 97.7 & 93.3 & 86.6 \\
10D & 20---80 & 41.5 & 34.7 & 12.9 & 10.5 \\
10D & 25---75 & 69.3 & 58.6 & 26.5 & 22.0 \\
10D & 30---70 & 95.4 & 88.2 & 47.9 & 40.7 \\
\end{longtable}}

Catalog union returns one admissible vector or prompt configuration. Strict tolerance is per-axis \textless10 and total \textless30. In 25--75, five-level union S coverage is 98.7\%, versus 88.0\% ordinary-L1 returned success; feasible-first retrieval uses the full existence coverage.

\subsection*{C.2 Single-Axis Additive Prediction and Retrieval}\label{c.2-single-axis-additive-prediction-and-retrieval}

Let \(\mathcal J\) contain evaluated joint nodes, excluding zero and single-axis configurations, and let \(Y(a)\) be their measured scores. Prediction MAE and target MAE are

\[\operatorname{MAE}_{\rm pred}=\frac{1}{|\mathcal J|d}\sum_{a\in\mathcal J}\sum_{j=1}^{d}|\widehat F_j(a)-Y_j(a)|,\]

\[\operatorname{MAE}_{\rm target}=\frac{1}{|\mathcal T|d}\sum_{t\in\mathcal T}\sum_{j=1}^{d}|Y_j(a_{\widehat F}(t))-t_j|.\]

Here \(a_{\widehat F}(t)\) is the configuration retrieved using estimated scores. The first error concerns a node's predicted response; the second concerns the measured response of the node returned for a target. Candidate sets, target sets, and retrieval rules are held equal within each comparison.

Score-additive prediction is \(\widehat s_j(a)=s_j(0)+\sum_k[s_j(a_ke_k)-s_j(0)]\). The clipped variant restricts the result to 0--100. With \(z_j(a)=\operatorname{logit}(\operatorname{clip}(s_j(a)/100,0.01,0.99))\), saturation-aware prediction is

\[\widehat s_j^{\rm logit}(a)=100\sigma\left(z_j(0)+\sum_k[z_j(a_ke_k)-z_j(0)]\right).\]

All cross-dimensional main effects are retained. Joint configurations are evaluation data only. Vector logit MAE is 5.244 over 1,836 admissible joint nodes, 4.968 over 1,225 nodes whose anchors are also admissible, and 2.701 over 190 nodes when coefficients are restricted to magnitude at most one. Corresponding prompt full/middle-three-level values are 3.973/3.169.

\textbf{Table C2. Returned-node success on 10,000 targets with common qualified candidates and feasible-first retrieval.} Additive retrieval is scored with archived joint responses. Intervals use 10,000 paired target-bootstrap resamples, conditional on the fixed catalogs.

{\small\begin{longtable}[]{@{}
  >{\raggedright\arraybackslash}p{(\linewidth - 8\tabcolsep) * \real{0.1667}}
  >{\raggedleft\arraybackslash}p{(\linewidth - 8\tabcolsep) * \real{0.2222}}
  >{\raggedleft\arraybackslash}p{(\linewidth - 8\tabcolsep) * \real{0.2222}}
  >{\raggedleft\arraybackslash}p{(\linewidth - 8\tabcolsep) * \real{0.2222}}
  >{\raggedright\arraybackslash}p{(\linewidth - 8\tabcolsep) * \real{0.1667}}@{}}
\toprule\noalign{}
\begin{minipage}[b]{\linewidth}\raggedright
Target domain
\end{minipage} & \begin{minipage}[b]{\linewidth}\raggedleft
Single-axis additive (\%)
\end{minipage} & \begin{minipage}[b]{\linewidth}\raggedleft
Joint grid (\%)
\end{minipage} & \begin{minipage}[b]{\linewidth}\raggedleft
Joint − additive
\end{minipage} & \begin{minipage}[b]{\linewidth}\raggedright
Paired-target 95\% CI
\end{minipage} \\
\midrule\noalign{}
\endhead
\bottomrule\noalign{}
\endlastfoot
\([25,75]^5\) & 62.22 & 91.85 & +29.63 & {[}28.73, 30.53{]} \\
\([20,80]^5\) & 50.95 & 80.56 & +29.61 & {[}28.72, 30.55{]} \\
\end{longtable}}

For composite targets in 25--75, mean actual L1 decreases from 44.16 to 29.98. With ordinary L1 for both methods, success is 60.49\% versus 76.09\%. Clipped addition preserves the primary success rate. Candidate and retrieval sensitivities are indexed in Appendix H.4.

\subsection*{C.3 Thinking-Mode Held-Out Repeats}\label{c.3-thinking-mode-held-out-repeats}

Two repeats estimate catalog and anchor responses, and the third evaluates them, rotating over three folds. Score prediction evaluates 72 non-single-axis nodes. Both retrieval methods use 81 candidates without filtering on held-out validity and the same 1,000 targets in 20--80.

\textbf{Table C3. All folds.}

{\small\begin{longtable}[]{@{}
  >{\raggedright\arraybackslash}p{(\linewidth - 10\tabcolsep) * \real{0.1304}}
  >{\raggedleft\arraybackslash}p{(\linewidth - 10\tabcolsep) * \real{0.1739}}
  >{\raggedleft\arraybackslash}p{(\linewidth - 10\tabcolsep) * \real{0.1739}}
  >{\raggedleft\arraybackslash}p{(\linewidth - 10\tabcolsep) * \real{0.1739}}
  >{\raggedleft\arraybackslash}p{(\linewidth - 10\tabcolsep) * \real{0.1739}}
  >{\raggedleft\arraybackslash}p{(\linewidth - 10\tabcolsep) * \real{0.1739}}@{}}
\toprule\noalign{}
\begin{minipage}[b]{\linewidth}\raggedright
Control
\end{minipage} & \begin{minipage}[b]{\linewidth}\raggedleft
Held-out repeat
\end{minipage} & \begin{minipage}[b]{\linewidth}\raggedleft
Logit prediction MAE
\end{minipage} & \begin{minipage}[b]{\linewidth}\raggedleft
Joint prediction MAE
\end{minipage} & \begin{minipage}[b]{\linewidth}\raggedleft
Logit target MAE
\end{minipage} & \begin{minipage}[b]{\linewidth}\raggedleft
Joint target MAE
\end{minipage} \\
\midrule\noalign{}
\endhead
\bottomrule\noalign{}
\endlastfoot
Vector & 1 & 9.123 & 5.684 & 15.490 & 16.158 \\
Vector & 2 & 7.874 & 5.695 & 18.514 & 15.578 \\
Vector & 3 & 7.797 & 5.403 & 16.411 & 15.676 \\
Prompt & 1 & 6.667 & 4.821 & 11.430 & 11.077 \\
Prompt & 2 & 7.206 & 4.843 & 12.001 & 11.231 \\
Prompt & 3 & 8.944 & 4.707 & 11.947 & 10.259 \\
\end{longtable}}

All six comparisons improve score prediction; five improve retrieval. Vector fold one has higher target error after joint calibration. Folds share items and overlapping training repeats, so their aggregate is a descriptive mean. The analysis indexed in Appendix H.4 also reports Q/B results.

\subsection*{C.4 Thinking-Mode Prompt Complementarity}\label{c.4-thinking-mode-prompt-complementarity}

Targets are the 2,401 combinations of 20/30/40/50/60/70/80 on four dimensions. Vector and prompt catalogs contain 81 configurations each. Union covers 63/529/1,311 targets at per-axis tolerances 5/10/15; adding 30 simultaneously combined configurations yields 63/534/1,345. No total-error constraint is added in this analysis.

The fixed-prompt endpoint study has 71 unique conditions, 384 responses each, totaling 27,264 generations. All-high prompt-only and positive-vector-combined validity counts are 376/384 and 373/384; all-low prompt-only and combined counts are 380/384 and 383/384. A separately selected high catalog configuration has 292/384 valid responses, 88 incomplete and four malformed; it is distinct from the all-positive-coefficient condition in the main table. Prompt low/middle/high S means increase in all 108 measured slices fixing the other dimensions. Appendix H.4 indexes the complete measurement records.

\subsection*{C.5 Disjoint-Bin Occupancy}\label{c.5-disjoint-bin-occupancy}

As a supplement to target-tolerance coverage, partition score space into disjoint cells and count cells containing at least one qualified measured node. For a fixed partition \(\mathcal P\),

\[N_{\rm bin}=\sum_{P\in\mathcal P}\mathbf1\{\exists a\in\mathcal G_{\rm adm}:\widehat F_\pi(a)\in P\},\qquad C_{\rm bin}=N_{\rm bin}/|\mathcal P|.\]

All coordinates must fall into the same joint cell. Intervals are left-closed/right-open, except that the final upper boundary is included; out-of-domain scores are not clipped. An occupied cell does not imply that every target inside it is reachable.

\textbf{Table C4. Qualified composite-score occupancy.} OFF boundaries are 25,35,45,55,65,75 on each axis. ON low/medium/high boundaries are 0,100/3,200/3,100, with independently calibrated catalogs.

{\small\begin{longtable}[]{@{}llrrr@{}}
\toprule\noalign{}
Mode & Catalog & Occupied & Total & Proportion \\
\midrule\noalign{}
\endhead
\bottomrule\noalign{}
\endlastfoot
OFF 5D & Three-level vector & 41 & 3,125 & 1.31\% \\
OFF 5D & Five-level vector & 342 & 3,125 & 10.94\% \\
OFF 5D & Nine-level vector & 845 & 3,125 & 27.04\% \\
ON 4D & Vector & 10 & 81 & 12.35\% \\
ON 4D & Prompt & 43 & 81 & 53.09\% \\
ON 4D & Union & 46 & 81 & 56.79\% \\
\end{longtable}}

\subsection*{C.6 Zero and Random-Configuration Baselines}\label{c.6-zero-and-random-configuration-baselines}

We reuse the same 1,000 targets in \([25,75]^5\), composite scores, and four-part filtering. The zero node retains the shared first-person LoRA but applies no vectors; its O/C/E/A/N scores are 64.45/61.31/50.81/69.41/34.31. A random singleton is drawn uniformly from all 1,854 qualified five-level nodes without inspecting the target. We compute the exact expectation over these nodes.

\textbf{Table C5. Single configurations versus catalog retrieval.} A realized singleton occupies zero or one in-domain cell; its fractional occupancy below is an expectation.

{\small\begin{longtable}[]{@{}
  >{\raggedright\arraybackslash}p{(\linewidth - 8\tabcolsep) * \real{0.1579}}
  >{\raggedleft\arraybackslash}p{(\linewidth - 8\tabcolsep) * \real{0.2105}}
  >{\raggedleft\arraybackslash}p{(\linewidth - 8\tabcolsep) * \real{0.2105}}
  >{\raggedleft\arraybackslash}p{(\linewidth - 8\tabcolsep) * \real{0.2105}}
  >{\raggedleft\arraybackslash}p{(\linewidth - 8\tabcolsep) * \real{0.2105}}@{}}
\toprule\noalign{}
\begin{minipage}[b]{\linewidth}\raggedright
Strategy
\end{minipage} & \begin{minipage}[b]{\linewidth}\raggedleft
Available nodes
\end{minipage} & \begin{minipage}[b]{\linewidth}\raggedleft
Broad coverage
\end{minipage} & \begin{minipage}[b]{\linewidth}\raggedleft
Strict coverage
\end{minipage} & \begin{minipage}[b]{\linewidth}\raggedleft
Occupied /3,125
\end{minipage} \\
\midrule\noalign{}
\endhead
\bottomrule\noalign{}
\endlastfoot
Zero vector & 1 & 10.30\% & 0.80\% & 1 \\
Uniform random singleton & 1 & 5.82\% & 0.31\% & 0.44 (expected) \\
Five-level catalog retrieval & 1,854 & 92.20\% & 44.00\% & 342 \\
\end{longtable}}

We also draw 243 or 3,125 configurations without replacement from all 59,049 measured nine-level configurations, then apply the same quality filter and retrieve from survivors. Each budget has 30 draws, seed 2026102608, without target-dependent sampling. This is retrospective analysis of measured records, not new GPU generation.

\textbf{Table C6. Equal-measurement-budget regular and random designs.} Random entries are mean±standard deviation across 30 subsets.

{\small\begin{longtable}[]{@{}
  >{\raggedright\arraybackslash}p{(\linewidth - 10\tabcolsep) * \real{0.1304}}
  >{\raggedleft\arraybackslash}p{(\linewidth - 10\tabcolsep) * \real{0.1739}}
  >{\raggedleft\arraybackslash}p{(\linewidth - 10\tabcolsep) * \real{0.1739}}
  >{\raggedleft\arraybackslash}p{(\linewidth - 10\tabcolsep) * \real{0.1739}}
  >{\raggedleft\arraybackslash}p{(\linewidth - 10\tabcolsep) * \real{0.1739}}
  >{\raggedleft\arraybackslash}p{(\linewidth - 10\tabcolsep) * \real{0.1739}}@{}}
\toprule\noalign{}
\begin{minipage}[b]{\linewidth}\raggedright
Design
\end{minipage} & \begin{minipage}[b]{\linewidth}\raggedleft
Measured
\end{minipage} & \begin{minipage}[b]{\linewidth}\raggedleft
Qualified
\end{minipage} & \begin{minipage}[b]{\linewidth}\raggedleft
Broad coverage
\end{minipage} & \begin{minipage}[b]{\linewidth}\raggedleft
Strict coverage
\end{minipage} & \begin{minipage}[b]{\linewidth}\raggedleft
Occupied bins
\end{minipage} \\
\midrule\noalign{}
\endhead
\bottomrule\noalign{}
\endlastfoot
Regular three-level & 243 & 120 & 71.20\% & 15.60\% & 41 \\
Random subset & 243 & 162.43±9.32 & 78.44±2.81\% & 21.55±1.34\% & 62.43±6.63 \\
Regular five-level & 3,125 & 1,854 & 92.20\% & 44.00\% & 342 \\
Random subset & 3,125 & 2,066.53±25.01 & 91.52±0.64\% & 42.93±0.58\% & 371.80±11.45 \\
\end{longtable}}

Random sampling improves coverage at the smaller budget and occupies more bins at both budgets; the regular five-level design has slightly higher tolerance coverage. Thus joint measurement and retrieval are the core contribution, while Cartesian enumeration is one sampling implementation rather than a universally optimal design. Source identities and all subsets' metrics are archived with the baseline analysis.

\subsection*{C.7 Complete Values Summarized in the Main Text}\label{c.7-complete-values-summarized-in-the-main-text}

\textbf{Table C7. Vector increments under identical prompts.} Each condition includes 384 responses, averaging three complete-generation repeats.

{\small\begin{longtable}[]{@{}
  >{\raggedright\arraybackslash}p{(\linewidth - 10\tabcolsep) * \real{0.1304}}
  >{\raggedleft\arraybackslash}p{(\linewidth - 10\tabcolsep) * \real{0.1739}}
  >{\raggedleft\arraybackslash}p{(\linewidth - 10\tabcolsep) * \real{0.1739}}
  >{\raggedleft\arraybackslash}p{(\linewidth - 10\tabcolsep) * \real{0.1739}}
  >{\raggedleft\arraybackslash}p{(\linewidth - 10\tabcolsep) * \real{0.1739}}
  >{\raggedleft\arraybackslash}p{(\linewidth - 10\tabcolsep) * \real{0.1739}}@{}}
\toprule\noalign{}
\begin{minipage}[b]{\linewidth}\raggedright
Endpoint/condition
\end{minipage} & \begin{minipage}[b]{\linewidth}\raggedleft
O
\end{minipage} & \begin{minipage}[b]{\linewidth}\raggedleft
R
\end{minipage} & \begin{minipage}[b]{\linewidth}\raggedleft
M
\end{minipage} & \begin{minipage}[b]{\linewidth}\raggedleft
A
\end{minipage} & \begin{minipage}[b]{\linewidth}\raggedleft
Answer validity
\end{minipage} \\
\midrule\noalign{}
\endhead
\bottomrule\noalign{}
\endlastfoot
High: prompt only & 80.10 & 79.79 & 83.05 & 80.76 & 97.92\% \\
High: vector + same prompt & \textbf{89.05} & \textbf{86.74} & \textbf{91.72} & \textbf{88.72} & 97.14\% \\
Low: prompt only & 11.81 & 8.63 & 16.84 & 27.08 & 98.96\% \\
Low: vector + same prompt & \textbf{4.51} & \textbf{4.89} & \textbf{8.51} & \textbf{25.17} & 99.74\% \\
\end{longtable}}

\textbf{Table C8. Prediction errors for non-thinking joint nodes.} Nodes and dimensions are equally weighted (protocol in C.2).

{\small\begin{longtable}[]{@{}
  >{\raggedright\arraybackslash}p{(\linewidth - 8\tabcolsep) * \real{0.1579}}
  >{\raggedleft\arraybackslash}p{(\linewidth - 8\tabcolsep) * \real{0.2105}}
  >{\raggedleft\arraybackslash}p{(\linewidth - 8\tabcolsep) * \real{0.2105}}
  >{\raggedleft\arraybackslash}p{(\linewidth - 8\tabcolsep) * \real{0.2105}}
  >{\raggedleft\arraybackslash}p{(\linewidth - 8\tabcolsep) * \real{0.2105}}@{}}
\toprule\noalign{}
\begin{minipage}[b]{\linewidth}\raggedright
Control
\end{minipage} & \begin{minipage}[b]{\linewidth}\raggedleft
Joint nodes
\end{minipage} & \begin{minipage}[b]{\linewidth}\raggedleft
Score-additive MAE
\end{minipage} & \begin{minipage}[b]{\linewidth}\raggedleft
Clipped MAE
\end{minipage} & \begin{minipage}[b]{\linewidth}\raggedleft
Logit additive MAE
\end{minipage} \\
\midrule\noalign{}
\endhead
\bottomrule\noalign{}
\endlastfoot
5D vector & 1,836 & 6.325 & 6.246 & \textbf{5.244} \\
5D prompt & 3,104 & 5.304 & 5.068 & \textbf{3.973} \\
\end{longtable}}

\section*{Appendix D Construction Ablations}\label{appendix-d-construction-ablations}

\subsection*{D.1 Cognitive--Behavioral Supervision Across Construction Pipelines}\label{d.1-cognitivebehavioral-supervision-across-construction-pipelines}

Holding the shared first-person adapter fixed, we compare questionnaire-only directions (FP-Q) and questionnaire-plus-behavior directions (FP-QB). Each selects its own layers and builds its five-level catalog; this compares supervision across complete construction pipelines. Section D.3 separately fixes layers.

\textbf{Table D1. Questionnaire and joint directions on the first-person model.} The same 10,000 targets in \([25,75]^5\) and tolerances are used; qualified columns apply all four filters.

{\small\begin{longtable}[]{@{}lrrrr@{}}
\toprule\noalign{}
Interface & FP-Q raw & FP-QB raw & FP-Q qualified & FP-QB qualified \\
\midrule\noalign{}
\endhead
\bottomrule\noalign{}
\endlastfoot
Questionnaire Q & 78.52\% & 81.55\% & 76.76\% & 77.32\% \\
Behavior B & 68.32\% & 89.42\% & 63.73\% & 80.81\% \\
Composite S & 87.41\% & 96.10\% & 86.38\% & 91.85\% \\
\end{longtable}}

\subsection*{D.2 Initialization, Matched-Data Extraction, and Refinement}\label{d.2-initialization-matched-data-extraction-and-refinement}

Fixed layers are O40/C32/E32/A16/N32. Original initialization uses final-position contrasts between high/low role prompts; comparisons retain its native norm or match it to the refined norm as specified in archived variants.

\textbf{Table D2. Original initialization and refined endpoint spans.} Each entry is the +2 minus −2 score.

{\small\begin{longtable}[]{@{}
  >{\raggedright\arraybackslash}p{(\linewidth - 12\tabcolsep) * \real{0.1111}}
  >{\raggedleft\arraybackslash}p{(\linewidth - 12\tabcolsep) * \real{0.1481}}
  >{\raggedleft\arraybackslash}p{(\linewidth - 12\tabcolsep) * \real{0.1481}}
  >{\raggedleft\arraybackslash}p{(\linewidth - 12\tabcolsep) * \real{0.1481}}
  >{\raggedleft\arraybackslash}p{(\linewidth - 12\tabcolsep) * \real{0.1481}}
  >{\raggedleft\arraybackslash}p{(\linewidth - 12\tabcolsep) * \real{0.1481}}
  >{\raggedleft\arraybackslash}p{(\linewidth - 12\tabcolsep) * \real{0.1481}}@{}}
\toprule\noalign{}
\begin{minipage}[b]{\linewidth}\raggedright
Dimension
\end{minipage} & \begin{minipage}[b]{\linewidth}\raggedleft
Initial Q span
\end{minipage} & \begin{minipage}[b]{\linewidth}\raggedleft
Refined Q span
\end{minipage} & \begin{minipage}[b]{\linewidth}\raggedleft
Initial B span
\end{minipage} & \begin{minipage}[b]{\linewidth}\raggedleft
Refined B span
\end{minipage} & \begin{minipage}[b]{\linewidth}\raggedleft
Initial S span
\end{minipage} & \begin{minipage}[b]{\linewidth}\raggedleft
Refined S span
\end{minipage} \\
\midrule\noalign{}
\endhead
\bottomrule\noalign{}
\endlastfoot
Openness (O) & 18.74 & 24.05 & 32.10 & 60.91 & 25.42 & 42.48 \\
Conscien\-tious\-ness (C) & 21.88 & 23.79 & 36.69 & 60.30 & 29.29 & 42.04 \\
Extraversion (E) & 21.43 & 27.78 & 49.24 & 83.60 & 35.34 & 55.69 \\
Agree\-able\-ness (A) & 0.71 & 16.79 & -3.74 & 30.03 & -1.51 & 23.41 \\
Neuroticism (N) & 24.66 & 35.63 & 27.15 & 41.65 & 25.91 & 38.64 \\
\end{longtable}}

Matched-data extraction uses every refinement target: 30 questionnaire items with two orders plus 16 behavioral pairs per dimension, totaling 380 pairs and 760 teacher-forced sequences. Q-mean and B-mean average content-token activation contrasts within each interface. QB-mean combines the group directions equally; QB-last uses the last target-content token. All four are norm-matched to the refined vector, with no parameter update.

\textbf{Table D3. All matched-data variants.} Layers, doses, and measurement interfaces are shared.

{\small\begin{longtable}[]{@{}llrrr@{}}
\toprule\noalign{}
Dimension & Direction & Q span & B span & S span \\
\midrule\noalign{}
\endhead
\bottomrule\noalign{}
\endlastfoot
O & Q-mean & 17.43 & 43.96 & 30.69 \\
O & B-mean & 8.95 & 40.72 & 24.84 \\
O & QB-mean & 17.09 & 52.89 & 34.99 \\
O & QB-last & 11.58 & 43.06 & 27.32 \\
O & Refined & 24.05 & 60.91 & 42.48 \\
C & Q-mean & 15.84 & 15.98 & 15.91 \\
C & B-mean & 3.55 & 39.84 & 21.70 \\
C & QB-mean & 15.31 & 36.77 & 26.04 \\
C & QB-last & 7.22 & 37.39 & 22.30 \\
C & Refined & 23.79 & 60.30 & 42.04 \\
E & Q-mean & 18.24 & 56.14 & 37.19 \\
E & B-mean & 11.79 & 63.33 & 37.56 \\
E & QB-mean & 15.57 & 70.41 & 42.99 \\
E & QB-last & 13.04 & 53.95 & 33.50 \\
E & Refined & 27.78 & 83.60 & 55.69 \\
A & Q-mean & 10.30 & -12.25 & -0.98 \\
A & B-mean & -12.32 & 35.35 & 11.52 \\
A & QB-mean & -10.82 & 35.19 & 12.18 \\
A & QB-last & -9.32 & 20.93 & 5.80 \\
A & Refined & 16.79 & 30.03 & 23.41 \\
N & Q-mean & 16.08 & 44.24 & 30.16 \\
N & B-mean & 12.45 & 48.70 & 30.58 \\
N & QB-mean & 18.81 & 56.25 & 37.53 \\
N & QB-last & 15.01 & 45.74 & 30.37 \\
N & Refined & 35.63 & 41.65 & 38.64 \\
\end{longtable}}

Refinement increases composite span relative to QB-mean on all axes and restores positive questionnaire direction for A. Direct extraction has larger B spans on A/N. Refined C at +2 has Q/B legal mass 97.69\%/98.19\%; the point is retained in curves and rejected by retrieval filtering. The corresponding measured own-interface masses for QB-mean exceed 99.6\%. Appendix H.4 indexes the per-dose values and source identities.

Matched-data extraction versus refinement has mean Q/B/S spans of 11.19/50.30/30.75 versus 25.61/55.30/40.45. Dimension-wise values already appear in the QB-mean and refined rows of Table D3.

\subsection*{D.3 Fixed-Layer Supervision}\label{d.3-fixed-layer-supervision}

Q-only, B-only, and QB share model, initialization, layers, 240 updates, and full-sequence injection, each producing 3,125 nodes. Q-only removes action loss; B-only removes choice and self-report losses; QB retains all components.

\textbf{Table D4. Full fixed-layer coverage.} Each domain has 10,000 targets, maximum per-axis error 20, and total error 60. Entries are unfiltered/qualified percentages.

{\small\begin{longtable}[]{@{}
  >{\raggedright\arraybackslash}p{(\linewidth - 8\tabcolsep) * \real{0.1667}}
  >{\raggedright\arraybackslash}p{(\linewidth - 8\tabcolsep) * \real{0.1667}}
  >{\raggedleft\arraybackslash}p{(\linewidth - 8\tabcolsep) * \real{0.2222}}
  >{\raggedleft\arraybackslash}p{(\linewidth - 8\tabcolsep) * \real{0.2222}}
  >{\raggedleft\arraybackslash}p{(\linewidth - 8\tabcolsep) * \real{0.2222}}@{}}
\toprule\noalign{}
\begin{minipage}[b]{\linewidth}\raggedright
Target domain
\end{minipage} & \begin{minipage}[b]{\linewidth}\raggedright
Refinement target
\end{minipage} & \begin{minipage}[b]{\linewidth}\raggedleft
Questionnaire Q
\end{minipage} & \begin{minipage}[b]{\linewidth}\raggedleft
Behavior B
\end{minipage} & \begin{minipage}[b]{\linewidth}\raggedleft
Composite S
\end{minipage} \\
\midrule\noalign{}
\endhead
\bottomrule\noalign{}
\endlastfoot
\([25,75]^5\) & Q & 82.55/78.76 & 73.11/65.75 & 87.90/85.64 \\
\([25,75]^5\) & B & 66.94/58.87 & 89.77/84.77 & 89.26/82.32 \\
\([25,75]^5\) & QB & 81.55/77.32 & 89.42/80.81 & 96.10/91.85 \\
\([20,80]^5\) & Q & 66.81/63.44 & 59.18/52.28 & 71.46/68.47 \\
\([20,80]^5\) & B & 41.04/35.26 & 72.95/66.29 & 68.70/61.82 \\
\([20,80]^5\) & QB & 65.73/61.55 & 79.97/70.16 & 87.02/80.56 \\
\end{longtable}}

This comparison evaluates how supervision changes reachability under both measurement interfaces. Independently selected pipelines in D.1 answer a different whole-system question.

\subsection*{D.4 Injection-Scope Strategies}\label{d.4-injection-scope-strategies}

Both strategies use the same earlier approved behavioral bank, questionnaires, initialization, budget, candidate layers, and development rule. This bank precedes later O/E revisions. Response-side injection covers the final prompt position and subsequent target positions; full-sequence injection additionally covers every nonpadding input position. Training and measurement use their respective scopes.

Independent selection yields O48/C32/E40/A40/N40 versus O32/C32/E32/A16/N32. The experiment therefore compares complete strategies including layer selection. First-token measurement reads the same final output position after different input-side intervention histories.

\textbf{Table D5. Endpoints and legal numeric mass under different input scopes.} All endpoints are retained.

{\small\begin{longtable}[]{@{}
  >{\raggedright\arraybackslash}p{(\linewidth - 8\tabcolsep) * \real{0.2000}}
  >{\raggedright\arraybackslash}p{(\linewidth - 8\tabcolsep) * \real{0.2000}}
  >{\raggedright\arraybackslash}p{(\linewidth - 8\tabcolsep) * \real{0.2000}}
  >{\raggedright\arraybackslash}p{(\linewidth - 8\tabcolsep) * \real{0.2000}}
  >{\raggedright\arraybackslash}p{(\linewidth - 8\tabcolsep) * \real{0.2000}}@{}}
\toprule\noalign{}
\begin{minipage}[b]{\linewidth}\raggedright
Dimension
\end{minipage} & \begin{minipage}[b]{\linewidth}\raggedright
Response-side Q (−2/0/+2)
\end{minipage} & \begin{minipage}[b]{\linewidth}\raggedright
Full-sequence Q (−2/0/+2)
\end{minipage} & \begin{minipage}[b]{\linewidth}\raggedright
Response-side legal mass (−2/+2)
\end{minipage} & \begin{minipage}[b]{\linewidth}\raggedright
Full-sequence legal mass (−2/+2)
\end{minipage} \\
\midrule\noalign{}
\endhead
\bottomrule\noalign{}
\endlastfoot
O & 52.24 / 52.95 / 53.45 & 37.60 / 52.95 / 61.60 & 99.79\% / 99.54\% & 97.72\% / 99.75\% \\
C & 57.17 / 57.89 / 58.31 & 44.37 / 57.89 / 68.25 & 99.76\% / 99.90\% & 99.96\% / 97.81\% \\
E & 47.61 / 48.53 / 51.62 & 29.19 / 48.53 / 64.82 & 99.90\% / 99.90\% & 99.96\% / 99.37\% \\
A & 60.10 / 70.44 / 71.35 & 61.35 / 70.44 / 77.99 & 99.90\% / 99.91\% & 99.84\% / 100.00\% \\
N & 50.59 / 51.08 / 72.37 & 39.56 / 51.08 / 75.21 & 99.88\% / 99.66\% & 99.93\% / 99.83\% \\
\end{longtable}}

Mean questionnaire span increases from 7.879 to 27.159. Quality filtering constrains deployment range. Appendix H.4 indexes the complete injection-scope measurements.

\subsection*{D.5 Density Effects by Measurement Component}\label{d.5-density-effects-by-measurement-component}

We retain the questionnaire and behavioral components separately to show how each contributes to the density results. Models, directions, score definitions, target sets, and filtering remain fixed.

\textbf{Table D6. Component coverage (\%) in 25--75.}

{\small\begin{longtable}[]{@{}lrrrr@{}}
\toprule\noalign{}
Levels/axis & Q broad & Q strict & B broad & B strict \\
\midrule\noalign{}
\endhead
\bottomrule\noalign{}
\endlastfoot
3 & 67.4 & 14.3 & 31.8 & 1.9 \\
5 & 75.3 & 23.4 & 80.6 & 24.1 \\
9 & \textbf{78.4} & \textbf{28.6} & \textbf{88.7} & \textbf{41.5} \\
\end{longtable}}

From five to nine levels, strict behavioral coverage increases by 17.4 percentage points and questionnaire coverage by 5.2. These component scores explain the composition of the unified measurement without changing the main retrieval rules.

\section*{Appendix E Probability Calibration and Complete Generation}\label{appendix-e-probability-calibration-and-complete-generation}

\subsection*{E.1 Five-Node Non-Thinking Ten-Dimensional Remeasurement}\label{e.1-five-node-non-thinking-ten-dimensional-remeasurement}

From 33,942 qualified nodes, zero is fixed in advance. Remaining nodes are ordered by composite L1 distance from zero, divided into four equal-rank strata, and sampled once per stratum with seed 20261014. Each node completes all 640 calibration presentations: 480 questionnaire and 160 behavioral prompts, sampled once each, totaling 3,200 outputs. Model, vectors, injection, and temperature 0.7 match the catalog; top-p is one, top-k zero, and the cap 8,192 tokens.

\textbf{Table E1. Configurations and remeasurement errors.} Coefficient order is O/C/E/A/N/R/T/V/M/Se.

{\small\begin{longtable}[]{@{}llrr@{}}
\toprule\noalign{}
Node & 10D coefficients & Dimension score MAE & Scoreable/640 \\
\midrule\noalign{}
\endhead
\bottomrule\noalign{}
\endlastfoot
29524 & 0,0,0,0,0,0,0,0,0,0 & 3.16 & 636 \\
24359 & 0,−1,1,−1,0,−1,1,−1,0,1 & 3.94 & 638 \\
42396 & 1,−1,0,0,−1,0,0,−1,1,−1 & 4.26 & 640 \\
21182 & 0,−1,−1,1,−1,−1,0,0,0,1 & 3.52 & 635 \\
1422 & −1,−1,−1,0,1,1,0,1,−1,−1 & 1.76 & 637 \\
\end{longtable}}

Questionnaire first lines must contain a legal digit in the current mapping; behavior requires legal A/B. A valid digit without a required rationale is scoreable but not fully instruction-compliant. Invalid answers remain missing, without zero imputation or resampling.

\textbf{Table E2. Scoreability and instruction compliance.}

{\small\begin{longtable}[]{@{}rrrr@{}}
\toprule\noalign{}
Generated & Scoreable answers & Scoreable rate & Fully compliant \\
\midrule\noalign{}
\endhead
\bottomrule\noalign{}
\endlastfoot
3,200 & 3,186 & 99.56\% & 3,026 \\
\end{longtable}}

\textbf{Table E3. Errors over 50 node--dimension cells.} Valid responses are aggregated within each cell using the fixed weights in Appendix B.3.

{\small\begin{longtable}[]{@{}
  >{\raggedleft\arraybackslash}p{(\linewidth - 10\tabcolsep) * \real{0.1739}}
  >{\raggedleft\arraybackslash}p{(\linewidth - 10\tabcolsep) * \real{0.1739}}
  >{\raggedleft\arraybackslash}p{(\linewidth - 10\tabcolsep) * \real{0.1739}}
  >{\raggedleft\arraybackslash}p{(\linewidth - 10\tabcolsep) * \real{0.1739}}
  >{\raggedleft\arraybackslash}p{(\linewidth - 10\tabcolsep) * \real{0.1739}}
  >{\raggedright\arraybackslash}p{(\linewidth - 10\tabcolsep) * \real{0.1304}}@{}}
\toprule\noalign{}
\begin{minipage}[b]{\linewidth}\raggedleft
Cells
\end{minipage} & \begin{minipage}[b]{\linewidth}\raggedleft
MAE
\end{minipage} & \begin{minipage}[b]{\linewidth}\raggedleft
Mean observed − predicted
\end{minipage} & \begin{minipage}[b]{\linewidth}\raggedleft
RMSE
\end{minipage} & \begin{minipage}[b]{\linewidth}\raggedleft
Max. absolute error
\end{minipage} & \begin{minipage}[b]{\linewidth}\raggedright
Five-node bootstrap MAE interval
\end{minipage} \\
\midrule\noalign{}
\endhead
\bottomrule\noalign{}
\endlastfoot
50 & 3.33 & −0.50 & 4.28 & 12.00 & 2.48---3.99 \\
\end{longtable}}

All responses terminate naturally without length truncation. Scoreability is 99.56\%; full instruction compliance is 94.56\%. Intervals use 2,000 resamples of the five complete nodes and describe between-node variation. This measures the deviation between probability expectations and a single round of actual answers for fixed configurations.

There are 14 missing answers. For planned count \(N\), valid count \(n\), and valid-score sum \(u\), the full-denominator mean lies in \([u/N,(u+100(N-n))/N]\). Aggregation with fixed scoring weights gives dimension-score MAE bounds of 3.19--3.51. The node-level scores and audit indexed in Appendix H.4 preserve all 50 prediction/response pairs, missing cells, and source identities.

\subsection*{E.2 Quality-Threshold Sensitivity}\label{e.2-quality-threshold-sensitivity}

On the same three-, five-, and nine-level non-thinking catalogs, we vary one threshold at a time while holding the other three fixed. No scores or model outputs are recomputed, so this directly evaluates how filtering changes the number of available candidates.

\textbf{Table E4. Qualified nodes after one-at-a-time threshold changes.} Full catalogs contain 243, 3,125, and 59,049 nodes. Original settings are legal mass \ensuremath{\geq}0.98, mapping-score difference \ensuremath{\leq}25, option-order probability difference \ensuremath{\leq}0.25, and label bias \ensuremath{\leq}0.90.

{\small\begin{longtable}[]{@{}lrrrr@{}}
\toprule\noalign{}
Criterion & Threshold & Three levels & Five levels & Nine levels \\
\midrule\noalign{}
\endhead
\bottomrule\noalign{}
\endlastfoot
Minimum legal mass & 0.90 & 134 & 2,008 & 40,924 \\
Minimum legal mass & 0.95 & 130 & 1,955 & 40,332 \\
Minimum legal mass & 0.98 & 120 & 1,854 & 38,932 \\
Minimum legal mass & 0.99 & 106 & 1,695 & 36,885 \\
Maximum mapping-score gap & 15 & 100 & 1,557 & 32,484 \\
Maximum mapping-score gap & 20 & 114 & 1,769 & 37,055 \\
Maximum mapping-score gap & 25 & 120 & 1,854 & 38,932 \\
Maximum mapping-score gap & 30 & 124 & 1,890 & 39,668 \\
Maximum order-probability gap & 0.15 & 61 & 892 & 18,620 \\
Maximum order-probability gap & 0.20 & 99 & 1,545 & 33,331 \\
Maximum order-probability gap & 0.25 & 120 & 1,854 & 38,932 \\
Maximum order-probability gap & 0.30 & 127 & 2,009 & 41,574 \\
Maximum label bias & 0.70 & 120 & 1,854 & 38,932 \\
Maximum label bias & 0.80 & 120 & 1,854 & 38,932 \\
Maximum label bias & 0.90 & 120 & 1,854 & 38,932 \\
Maximum label bias & 0.95 & 120 & 1,854 & 38,932 \\
\end{longtable}}

Within the examined ranges, order consistency has the largest effect on candidate count. Label-bias thresholds do not further change the admissible set when the other three constraints are fixed, indicating overlapping exclusions. Tightening legal mass from 98\% to 99\% still retains 36,885 nine-level nodes; occupancy of 25--75 score bins decreases from 845 to 831. Applications can select thresholds according to acceptable output risk and required candidate count.

These statistics concern measured probabilities and consistency, not complete-dialogue failure rates. Independently sampling new nodes to compare complete-generation validity inside and outside the admissible set, and varying scoring temperature, require separate experiments.

\section*{Appendix F External Scenarios and Interactive Applications}\label{appendix-f-external-scenarios-and-interactive-applications}

\subsection*{F.1 Chinese Situational-Judgment Items}\label{f.1-chinese-situational-judgment-items}

We use all 40 Study 3 items and original binary keys from Li et al.~\citep{ref25}. Five narrow facets have eight items each: openness to ideas, self-discipline, gregariousness, compliance/conflict handling, and self-consciousness/negative-evaluation sensitivity. Four cyclic option orders are used, and keys are absent from prompts. For each supervision arm, distinct nodes are selected at the 10th/50th/90th percentiles of calibrated Q; other-axis distance from 50 is only a secondary ranking criterion, not a hard constraint.

Forty-five nonzero configurations plus the shared zero baseline yield 7,360 unconstrained greedy, non-thinking responses with a 32-token cap. The 160 baseline outputs are reused after identity checks. Each designated facet has 1,440 valid outputs; 30 invalid outputs occur on background facets. Observed scores follow selected-option keys; conditional scores average first-token legal-option probabilities.

\textbf{Table F1. Calibration coordinates and external responses.} Triples denote low/middle/high.

{\small\begin{longtable}[]{@{}
  >{\raggedright\arraybackslash}p{(\linewidth - 8\tabcolsep) * \real{0.2000}}
  >{\raggedright\arraybackslash}p{(\linewidth - 8\tabcolsep) * \real{0.2000}}
  >{\raggedright\arraybackslash}p{(\linewidth - 8\tabcolsep) * \real{0.2000}}
  >{\raggedright\arraybackslash}p{(\linewidth - 8\tabcolsep) * \real{0.2000}}
  >{\raggedright\arraybackslash}p{(\linewidth - 8\tabcolsep) * \real{0.2000}}@{}}
\toprule\noalign{}
\begin{minipage}[b]{\linewidth}\raggedright
Training arm
\end{minipage} & \begin{minipage}[b]{\linewidth}\raggedright
Facet
\end{minipage} & \begin{minipage}[b]{\linewidth}\raggedright
Calibration Q
\end{minipage} & \begin{minipage}[b]{\linewidth}\raggedright
Observed keyed score
\end{minipage} & \begin{minipage}[b]{\linewidth}\raggedright
Conditional score
\end{minipage} \\
\midrule\noalign{}
\endhead
\bottomrule\noalign{}
\endlastfoot
Q & O & 43.80/56.00/66.73 & 96.88/100.00/100.00 & 86.78/98.41/99.34 \\
Q & C & 36.08/55.50/75.96 & 93.75/100.00/100.00 & 91.21/96.23/97.05 \\
Q & E & 37.07/56.57/69.77 & 6.25/31.25/31.25 & 11.93/36.10/39.51 \\
Q & A & 43.39/61.31/78.53 & 93.75/93.75/90.63 & 86.27/91.53/90.48 \\
Q & N & 32.58/59.44/77.58 & 3.13/0.00/9.38 & 8.77/7.11/13.08 \\
B & O & 50.89/53.90/57.14 & 100.00/100.00/100.00 & 98.33/92.32/98.50 \\
B & C & 50.93/55.23/59.69 & 100.00/100.00/100.00 & 95.50/94.58/93.05 \\
B & E & 47.27/53.73/59.82 & 21.88/12.50/28.13 & 21.46/21.23/38.01 \\
B & A & 55.06/62.64/68.01 & 87.50/81.25/87.50 & 81.58/75.75/84.38 \\
B & N & 52.65/57.85/65.16 & 0.00/0.00/0.00 & 4.76/10.72/11.69 \\
QB & O & 41.45/57.08/64.61 & 93.75/100.00/100.00 & 84.56/99.49/99.20 \\
QB & C & 39.00/57.19/70.39 & 100.00/100.00/100.00 & 94.68/94.24/91.24 \\
QB & E & 34.35/57.69/69.98 & 6.25/25.00/46.88 & 10.28/28.58/50.42 \\
QB & A & 48.61/60.98/72.19 & 62.50/84.38/84.38 & 58.07/78.50/83.14 \\
QB & N & 40.71/55.82/75.55 & 0.00/0.00/3.13 & 4.74/11.87/11.76 \\
\end{longtable}}

\textbf{Table F2. Adjacent-level directions.} Increases/ties/decreases score 1/0.5/0 across ten adjacent facet-mean comparisons.

{\small\begin{longtable}[]{@{}
  >{\raggedright\arraybackslash}p{(\linewidth - 10\tabcolsep) * \real{0.1304}}
  >{\raggedleft\arraybackslash}p{(\linewidth - 10\tabcolsep) * \real{0.1739}}
  >{\raggedleft\arraybackslash}p{(\linewidth - 10\tabcolsep) * \real{0.1739}}
  >{\raggedleft\arraybackslash}p{(\linewidth - 10\tabcolsep) * \real{0.1739}}
  >{\raggedleft\arraybackslash}p{(\linewidth - 10\tabcolsep) * \real{0.1739}}
  >{\raggedleft\arraybackslash}p{(\linewidth - 10\tabcolsep) * \real{0.1739}}@{}}
\toprule\noalign{}
\begin{minipage}[b]{\linewidth}\raggedright
Training arm
\end{minipage} & \begin{minipage}[b]{\linewidth}\raggedleft
Facet means: up/tie/down
\end{minipage} & \begin{minipage}[b]{\linewidth}\raggedleft
Direction score \(M_r\)
\end{minipage} & \begin{minipage}[b]{\linewidth}\raggedleft
Nondecrease \(D_r\)
\end{minipage} & \begin{minipage}[b]{\linewidth}\raggedleft
Item-level: up/tie/down
\end{minipage} & \begin{minipage}[b]{\linewidth}\raggedleft
Item nondecrease
\end{minipage} \\
\midrule\noalign{}
\endhead
\bottomrule\noalign{}
\endlastfoot
Q & 4/4/2 & 60\% & 8/10 (80\%) & 7/70/3 & 77/80 (96.25\%) \\
B & 2/6/2 & 50\% & 8/10 (80\%) & 5/71/4 & 76/80 (95.00\%) \\
QB & 5/5/0 & 75\% & 10/10 (100\%) & 14/65/1 & 79/80 (98.75\%) \\
\end{longtable}}

Joint-supervision observed choices show five increases and five ties, with clearer differentiation on E/A. Conditional probabilities show six increases and four decreases, retaining variation hidden by discrete choices. Shared baseline O/C/E/A/N scores are 100/100/21.88/87.50/3.13. Per-item records retain invalid denominators, background responses, and option orders.

\subsection*{F.2 Early Open-Generation Diagnosis}\label{f.2-early-open-generation-diagnosis}

Before quality filtering, original-model questionnaire, first-person questionnaire, and first-person joint pipelines generated 300 thinking-mode open answers. Naturally ended, complete outputs among 100 per arm were 62/92/53. Across 20 vector-only pairs per arm, positive/tied/reversed/unjudgeable differences were 1/9/2/8, 2/6/0/12, and 3/2/0/15. This diagnosis motivated separate quality checks for open outputs. Raw responses and archived labels remain available; primary external analyses use the frozen protocols in F.1.

\subsection*{F.3 Chinese PersonalityBench Generation and Blind Judging}\label{f.3-chinese-personalitybench-generation-and-blind-judging}

Section 5.1 uses Qwen3.5-27B with the same first-person LoRA, thinking disabled, greedy decoding, a 512-new-token cap, and repetition penalty 1.0. The system instruction remains ``You are a person; answer according to the facts of the scenario.'' Composite-score levels select nodes but are not written into personality prompts. Fifteen dimension--level configurations each answer all Chinese items for their focal dimension once. No node is reselected based on response quality.

Judging uses the official DeepSeek API, requested model \texttt{deepseek-flash}, thinking disabled, temperature zero, and JSON output with a 1,800-token cap. Inputs contain only the dimension, scenario question, and raw response. Ratings 1/2/3/4/5 mean clearly low, somewhat low, genuinely intermediate or mixed, somewhat high, and clearly high. Missing trait evidence is marked missing rather than assigned three. Ratings prioritize expressed choices, intended actions, preferences, and emotions; self-labels and politeness alone do not establish trait expression. Paired endpoint differences use only same-item pairs with both responses scoreable.

The Chinese items are this study's translation, not an official Chinese release, and retain all 459 source identifiers. The archive preserves translations, manual adjustments, selected profiles, the rubric, original responses, API-returned model identifiers, and individual scores. Table 4 is aggregated from these records. Source and protocol records are indexed in Appendix H.4.

\section*{Appendix G Cross-Dimensional Structure and Population Reference}\label{appendix-g-cross-dimensional-structure-and-population-reference}

\subsection*{G.1 Joint-Catalog Correlations}\label{g.1-joint-catalog-correlations}

Human trait correlations have been studied through higher-order factors and common rating effects \citep{ref39,ref40}. We use one population sample as a sign reference, not as a causal model of internal interventions.

All3,125 unfiltered five-dimensional configurations are equally weighted to compute Pearson correlation signs of the composite dimension scores. The population reference is the lower triangle of Table3, Study1, in Føllesdal and Soto's Norwegian BFI-2 study, with pairwise sample sizes599--601 \citep{ref41}.

\textbf{Table G1. Signs for ten unordered dimension pairs.}

{\small\begin{longtable}[]{@{}lll@{}}
\toprule\noalign{}
Dimension pair & Population sign & Model-score sign \\
\midrule\noalign{}
\endhead
\bottomrule\noalign{}
\endlastfoot
O---C & + (near zero) & + \\
O---E & + & + \\
O---A & + & + \\
O---N & − (weak) & − \\
C---E & + & + \\
C---A & + & + \\
C---N & − & − \\
E---A & + & + \\
E---N & − & − \\
A---N & − & − \\
\end{longtable}}

All ten composite dimension-score correlation signs agree with the population point-estimate signs. The catalog and population sample thus exhibit similar cross-dimensional association patterns, providing a descriptive reference for interpreting joint coordinates.

\subsection*{G.2 Directed Single-Axis Responses before and after Refinement}\label{g.2-directed-single-axis-responses-before-and-after-refinement}

Fix other coefficients at zero and compute \(D_{jk}(h)=[F_j(he_k)-F_j(-he_k)]/(2h)\). Twenty directed off-diagonal effects are compared with the corresponding population signs at \(h=1,2\), holding layers fixed between initialization and refinement.

\textbf{Table G2. Directed-response sign agreement.}

{\small\begin{longtable}[]{@{}
  >{\raggedright\arraybackslash}p{(\linewidth - 10\tabcolsep) * \real{0.1364}}
  >{\raggedright\arraybackslash}p{(\linewidth - 10\tabcolsep) * \real{0.1364}}
  >{\raggedleft\arraybackslash}p{(\linewidth - 10\tabcolsep) * \real{0.1818}}
  >{\raggedleft\arraybackslash}p{(\linewidth - 10\tabcolsep) * \real{0.1818}}
  >{\raggedleft\arraybackslash}p{(\linewidth - 10\tabcolsep) * \real{0.1818}}
  >{\raggedleft\arraybackslash}p{(\linewidth - 10\tabcolsep) * \real{0.1818}}@{}}
\toprule\noalign{}
\begin{minipage}[b]{\linewidth}\raggedright
Half-width h
\end{minipage} & \begin{minipage}[b]{\linewidth}\raggedright
Interface
\end{minipage} & \begin{minipage}[b]{\linewidth}\raggedleft
Initial /20
\end{minipage} & \begin{minipage}[b]{\linewidth}\raggedleft
Refined /20
\end{minipage} & \begin{minipage}[b]{\linewidth}\raggedleft
Initial /16, weak excluded
\end{minipage} & \begin{minipage}[b]{\linewidth}\raggedleft
Refined /16, weak excluded
\end{minipage} \\
\midrule\noalign{}
\endhead
\bottomrule\noalign{}
\endlastfoot
1 & Questionnaire score Q & 16 & 18 & 14 & 15 \\
1 & Behavior score B & 17 & 13 & 13 & 10 \\
1 & Composite score S & 17 & 16 & 13 & 13 \\
2 & Questionnaire score Q & 17 & 19 & 14 & 16 \\
2 & Behavior score B & 17 & 13 & 13 & 10 \\
2 & Composite score S & 17 & 15 & 13 & 12 \\
\end{longtable}}

Questionnaire-score agreement rises from16/20 to18/20 at \(h=1\), and17/20 to19/20 at \(h=2\). TableG2 explicitly distinguishes questionnaire scores \(Q\), behavioral scores \(B\), and composite scores \(S\): behavioral agreement changes from17/20 to13/20; composite agreement changes from17/20 to16/20 or15/20. TableG1 reports only composite-score correlations. Directed intervention effects locate spillovers from individual directions; catalog correlations describe the overall configuration geometry. AppendixH.4 indexes the complete matrices and raw-response identities.

\section*{Appendix H Execution, Timing, and Evidence}\label{appendix-h-execution-timing-and-evidence}

\subsection*{H.1 Per-Token Routing and Scheduling for Complete Generation}\label{h.1-per-token-routing-and-scheduling-for-complete-generation}

The six-dimensional thinking instance uses independent single-GPU vLLM replicas. Each replica loads the same full model and node-vector bank, with at most 32 active sequences and a 2,048-token prefill budget per scheduling step. Output and total-context limits are 4,096 and 32,768 tokens. The model uses BF16, with an FP32 output head and vector bank; no LoRA is loaded and CUDA Graph is disabled. Explicit mappings convert node identifiers to bank indices; node identifiers are not array offsets.

\textbf{Algorithm H1. Continuous generation and calibration with mixed configurations.}

\begin{Verbatim}[breaklines=true,fontsize=\small]
Inputs: frozen model M_FP, directions V, configurations G,
        items and repeats X, protocol pi, replica count P
Output: response catalog C_pi
1  B <- FP32SumByLayer(V, G); p <- BuildPersonaIndex(G)
2  W <- EnumerateRequests(G, X, pi)  # node, prompt, seed, budget
3  parallel for replica r = 1, ..., P:
4      LoadOnce(M_FP, B); CheckMixedAndReorderedRouting()
5      while unfinished work remains:
6          AdmitUncommittedRequests(W, atomic_claim=True)
7          U <- ScheduleChunkedPrefillAndDecode()
8          route <- RebuildTokenOwnersAfterReorder(U, p)
9          key <- BindCacheToModelVectorsProtocolAndPersona(U)
10         Forward(U, key, layer_injection=B[layer, route])
11         SampleEachRequestUsingItsOwnParameters()
12         for each completed request:
13             AtomicCommit(raw_thought, final, finish, seed, digest)
14             ReleaseSlotAndAdmitNext()
15 return ParseFinalsAndAggregate(all_committed, pi)
\end{Verbatim}

The engine manages paged caches, whose identities additionally bind the model, vector bank, and profile. Different configurations do not share post-injection states. Startup checks include A/B/A profiles in one batch and reversed request order to detect routing errors after scheduler reordering. Raw reasoning, final answers, completion status, and invalid outputs are all saved before scoring. Recovery fills only unfinished work; uncertain commits are reconciled rather than blindly replayed.

With 16 GPUs, demonstration generation through final verification takes 4 h 11 min 18 s. Formal measurement of 559,872 complete outputs takes 3 h 5 min 4 s, averaging 50.42 answers/s overall and 3.15 per GPU. This stage includes startup overhead, excludes the preceding 54 min 28 s of full-parameter post-training, and is not a paired speedup against the older first-token engine. The chat service reuses token routing with a 16,384-token context and lower concurrency limits, without altering catalog scores.

\subsection*{H.2 Boundary Reuse and Cost in Earlier First-Token Measurement}\label{h.2-boundary-reuse-and-cost-in-earlier-first-token-measurement}

Earlier non-thinking experiments reuse states according to intervention dependencies. Computation before the first injection is coefficient-independent; later states depend only on upstream coefficients that have already acted, delaying branches for deeper coefficients. Boundary-cache keys bind model, adapter, item tokens, positions, masks, vectors, and upstream coefficients, and retain complete continuation states. This differs from the persona-isolated prefix caching in Appendix H.1.

Let zero-indexed injection layers be \(b_1<\cdots<b_K\), with \(g_s\) combinations at layer group \(s\) and \(P_s=\prod_{u\le s}g_u\) upstream combinations. Ideal boundary reuse has block-execution counts

\[\begin{aligned}
C_{\rm ind}&=N_GL,\\
C_{\rm reuse}&=(b_1+1)+\sum_{s=1}^{K-1}P_s(b_{s+1}-b_s)\\
&\quad+N_G(L-1-b_K).
\end{aligned}\]

This count explains frozen-forward savings from delayed branching, not complete autoregressive generation time. Original checks cover zero vectors, signs, masks, padding, causality, and cached decoding. Option log-probability and probability tolerances are \(10^{-4}\) and \(10^{-5}\), with separate cross-replica anchor checks.

The ten-dimensional design uses 640 first-token probes per node, totaling 37,791,360 probes for 59,049 nodes. One shared adapter and \(dH\) direction parameters serve all profiles; catalog storage grows linearly with measured nodes. Full Cartesian enumeration remains multiplicative in dimension count. Acceleration removes repeated frozen computation rather than training a separate model for each configuration.

The original ten-dimensional catalog takes approximately 13.56 hours from admission to completion, using at most 16 replicas. Its five-dimensional reference reuses 1,800 nodes and measures 1,325 more in approximately 40.6 minutes. An early alternative vLLM implementation did not meet the then-current numerical and speed criteria, so those first-token results use the original pipeline. The new complete-generation implementation in Appendix H.1 builds a separate six-dimensional catalog and does not replace historical measurements.

\subsection*{H.3 Density-Study Timing and Unfiltered Results}\label{h.3-density-study-timing-and-unfiltered-results}

The nine-level catalog contains 59,049 nodes: 3,125 reused integer nodes, 14,183 previously completed half-step nodes, and 41,741 newly measured nodes. Sixteen replicas complete supplementation and aggregation in approximately ten hours after numerical checks and paired benchmarking. This is not a from-scratch runtime for all nine-level nodes.

\textbf{Table H1. Paired same-device acceleration.} Each host repeats three comparisons of three full nodes per path. Batch size 32, precision, and SDPA/FLA operators are unchanged. Reference execution disables cross-node boundary-state reuse, not all optimizations.

{\small\begin{longtable}[]{@{}
  >{\raggedright\arraybackslash}p{(\linewidth - 10\tabcolsep) * \real{0.1304}}
  >{\raggedleft\arraybackslash}p{(\linewidth - 10\tabcolsep) * \real{0.1739}}
  >{\raggedleft\arraybackslash}p{(\linewidth - 10\tabcolsep) * \real{0.1739}}
  >{\raggedleft\arraybackslash}p{(\linewidth - 10\tabcolsep) * \real{0.1739}}
  >{\raggedleft\arraybackslash}p{(\linewidth - 10\tabcolsep) * \real{0.1739}}
  >{\raggedleft\arraybackslash}p{(\linewidth - 10\tabcolsep) * \real{0.1739}}@{}}
\toprule\noalign{}
\begin{minipage}[b]{\linewidth}\raggedright
Host
\end{minipage} & \begin{minipage}[b]{\linewidth}\raggedleft
Reference s/node
\end{minipage} & \begin{minipage}[b]{\linewidth}\raggedleft
Fill-inclusive s/node
\end{minipage} & \begin{minipage}[b]{\linewidth}\raggedleft
Warm s/node
\end{minipage} & \begin{minipage}[b]{\linewidth}\raggedleft
Fill-inclusive speedup
\end{minipage} & \begin{minipage}[b]{\linewidth}\raggedleft
Warm speedup
\end{minipage} \\
\midrule\noalign{}
\endhead
\bottomrule\noalign{}
\endlastfoot
Host group 1 & 35.592 & 20.497 & 12.880 & 1.736 & 2.763 \\
Host group 2 & 34.563 & 19.879 & 12.492 & 1.739 & 2.767 \\
\end{longtable}}

Maximum paired differences in option probabilities, log probabilities, and legal mass are zero on both hosts. Observed formal throughput is approximately 24 first-token probes per GPU per second, depending on cache reuse and node grouping. Parallelism is distinct from the same-device algorithm speedup in Table H1.

Without filtering, three/five/nine-level composite broad coverage in 25--75 is 92.3\%/95.3\%/96.4\%; strict coverage is 31.4\%/55.5\%/64.1\%. Table 3 reports qualified coverage. Retaining both separates raw reachability from usable regions under quality constraints. Table 3 reports density outcomes; Table H1 reports paired timing.

\subsection*{H.4 Evidence Organization}\label{h.4-evidence-organization}

The evidence chain is protocol and inputs → raw commits → analysis → paper tables. Supporting files retain item identities, source digests, repeated-generation denominators, missing responses, background coordinates, and quality. The source index below identifies the main catalogs, component studies, and this revision. Scores remain response coordinates tied to the specified evaluation protocol.

Provenance record identifiers (retained by the authors; not bundled with the typesetting source):

{\small\begin{longtable}[]{@{}
  >{\raggedright\arraybackslash}p{(\linewidth - 2\tabcolsep) * \real{0.5000}}
  >{\raggedright\arraybackslash}p{(\linewidth - 2\tabcolsep) * \real{0.5000}}@{}}
\toprule\noalign{}
\begin{minipage}[b]{\linewidth}\raggedright
Study
\end{minipage} & \begin{minipage}[b]{\linewidth}\raggedright
Supporting record
\end{minipage} \\
\midrule\noalign{}
\endhead
\bottomrule\noalign{}
\endlastfoot
Templates and splits & \nolinkurl{evidence/current_prompt_examples.json} \\
Main catalogs and complementarity & \nolinkurl{evidence/completed_catalog_coverage.json}; \nolinkurl{evidence/v25_extension_verified.json} \\
Six-dimensional full thinking workflow & \nolinkurl{evidence/huihui_six_pipeline_20261104.json} \\
Fixed-layer comparisons and retrieval & \nolinkurl{evidence/fixed_layer_coverage_extract.json}; \nolinkurl{evidence/joint_vs_additive_summary.json} \\
Matched-data refinement & \nolinkurl{evidence/same_data_refinement_20261025_curves.csv}; \nolinkurl{evidence/same_data_refinement_20261025_provenance.json} \\
Injection scope & \nolinkurl{evidence/injection_scope_20261008_summary.json} \\
Joint prediction and held-out repeats & \nolinkurl{evidence/joint_additivity_20261025_results.json} \\
Ten-dimensional non-thinking repeats & \nolinkurl{evidence/tenaxis_off5_node_scores.csv}; \nolinkurl{evidence/tenaxis_off5_independent_audit.json} \\
Cross-axis directions & \nolinkurl{evidence/directional_refinement_20261025.json}; \nolinkurl{evidence/directional_grid_correlation_20261025.json} \\
Density and timing & \nolinkurl{evidence/density_complete.json}; \nolinkurl{evidence/density_benchmark_summary.json} \\
Score bins and threshold sensitivity & \nolinkurl{evidence/static_bins_20261026.json} \\
Six-method HEXACO comparison & \nolinkurl{evidence/hexaco_six_methods_joint_coverage.json}; \nolinkurl{evidence/hexaco_six_methods_joint_nodes.json}; \nolinkurl{evidence/hexaco_six_methods_range_records.json} \\
Market interaction case & \nolinkurl{evidence/seller_market_case_20261027.json} \\
Revision audit & \nolinkurl{evidence/revision_v28.json} \\
Chinese PersonalityBench & \nolinkurl{evidence/pb_scene_revision_20261027.json}; \nolinkurl{evidence/pb_reproduction_20261027.json} \\
Main-aligned HEXACO ranges & \nolinkurl{evidence/hexaco_mainaligned_range_analysis.json} \\
Main-aligned HEXACO joint results & \nolinkurl{evidence/hexaco_mainaligned_joint_coverage.json}; \nolinkurl{evidence/hexaco_mainaligned_joint_nodes.json}; \nolinkurl{evidence/hexaco_mainaligned_verified.json} \\
\end{longtable}}

\section*{Appendix I Public Six-Dimensional Inventory: Construction, Range, and Joint Measurement}\label{appendix-i-public-six-dimensional-inventory-construction-range-and-joint-measurement}

\subsection*{I.1 Direction Construction and Complete-Response Scoring}\label{i.1-direction-construction-and-complete-response-scoring}

The IPIP HEXACO inventory is split by 24 facets, each contributing six training, two development, and two test items. Each dimension has eight test items and three complete-response samples under each of two numeric mappings. Thinking is disabled; temperature is 0.7, top-p is one, top-k is zero, the cap is 192 new tokens, and batch size is 16. Selected layers, a relative reference-norm budget of 0.1, and full-input/decode injection remain fixed during evaluation.

Figure 1 presents the fixed-range curves. Original English items and keys are from the \href{https://ipip.ori.org/newHEXACO_PI_key.htm}{IPIP public inventory}, based on Ashton, Lee, and Goldberg (2007; \href{https://ipip.ori.org/newHEXACO_PI_table.htm}{bibliographic information}). HEXACO E denotes emotionality and X denotes extraversion, distinct from the Big Five abbreviations.

CAA, the independent linear probe, role-endpoint initialization, and refinement use the same training items. CAA averages high-minus-low activation contrasts at answer positions; the independent linear probe fits a classification direction to those high/low activations. Role-endpoint initialization prefixes each questionnaire prompt with a high or low trait description, keeping facts, questions, and encoding unchanged. It averages the high-minus-low contrasts at the last valid prompt position. Reference scale is the median L2 norm at the last position of role-free training prompts. Refinement optimizes this initialization for 240 steps with choice loss, complete self-report loss, neutral preservation, and norm regularization, without behavioral training items. PERSONA uses twenty training items per dimension and five elicitation/suppression prompt pairs to generate 2,400 construction responses. Official DeepSeek numeric ratings filter trait expression and coherence; content-mean activation contrasts then construct independent high and low poles.

Candidate outputs are decoder blocks 8, 16, 24, 32, 40, 48, 56, and 60, numbered from one. All methods except PERSONA select layers independently per dimension on development data; PERSONA selects one common layer across twelve poles. Directly constructed directions have fixed length equal to 0.1 times the reference scale. Refinement and BiPO project into the same norm ball after every update and retain their learned lengths without renormalization; this matches norm budgets, not exact final lengths. The independent probe represents a single-axis classification direction, not SAS's sequential multi-axis augmentation. PERSONA uses inventory-response development selection, matched norms, and full-sequence injection. These comparisons adapt construction methods to shared data interfaces, norm budgets, and injection rules.

BiPO \citep{ref47} initializes the vector at zero and applies a randomly sampled positive or negative coefficient on each update. Its bidirectional DPO objective uses summed log probabilities of complete high/low self-report responses, including the end token, relative to the frozen unsteered model. Both A/B mappings receive equal weight. AdamW uses weight decay 0.05, gradient clipping at one, and 240 updates, without our auxiliary choice, self-report cross-entropy, or neutral-preservation terms. We parameterize \(v=0.1ru\) and project \(u\) into the unit L2 ball. Development selection considers \(\beta\in\{0.1,0.5\}\) and constant learning rates \(\{0.001,0.003\}\) for dimensionless \(u\) at each of the same eight layers, giving 32 candidates per dimension. BiPO has four hyperparameter settings per layer versus one for refinement. This matched-data, norm-budget protocol retains BiPO's zero initialization and bidirectional objective, but uses the shared 240-update budget rather than its original dataset and 100-epoch schedule.

PERSONA's negative half-axis uses its independent low-pole direction with positive strength; its positive half-axis uses the high pole. Both share the unsteered origin. Complete-response means are computed within item--mapping cells and then weighted equally. Unparseable answers are not imputed as neutral scores.

\subsection*{I.2 Adaptive Range and Response Validity}\label{i.2-adaptive-range-and-response-validity}

Six methods, six dimensions, and two signs yield 72 branches. Coarse absolute strengths are 0.5, 1, 1.5, 2, 3, 4, 6, 8, 10, 12, 16, 24, 32, 48, 64, and 100. Identical existing conditions are reused. At the first outward failure below 90\% validity, the last passing/first failing interval is bisected to width at most 0.25. At least 44 of 48 answers must pass. All branches encountered failure before reaching 100.

Valid answers terminate normally, begin with a legal 1--5 digit, include a brief explanation, and contain no thinking tags. The check does not include human semantic-coherence judgments. Means are computed within item--mapping cells before equal-weight aggregation. Invalid texts are preserved, not assigned 50. Usable span is the maximum minus minimum score among measured conditions that pass before the first outward failure; connecting curves does not establish behavior at unmeasured coefficients.

\begin{figure}
\centering
\pandocbounded{\includegraphics[keepaspectratio,alt={Figure I1. Expanded complete-response score curves. Dots meet the initial 48-response validity threshold of at least 90\%; crosses do not.}]{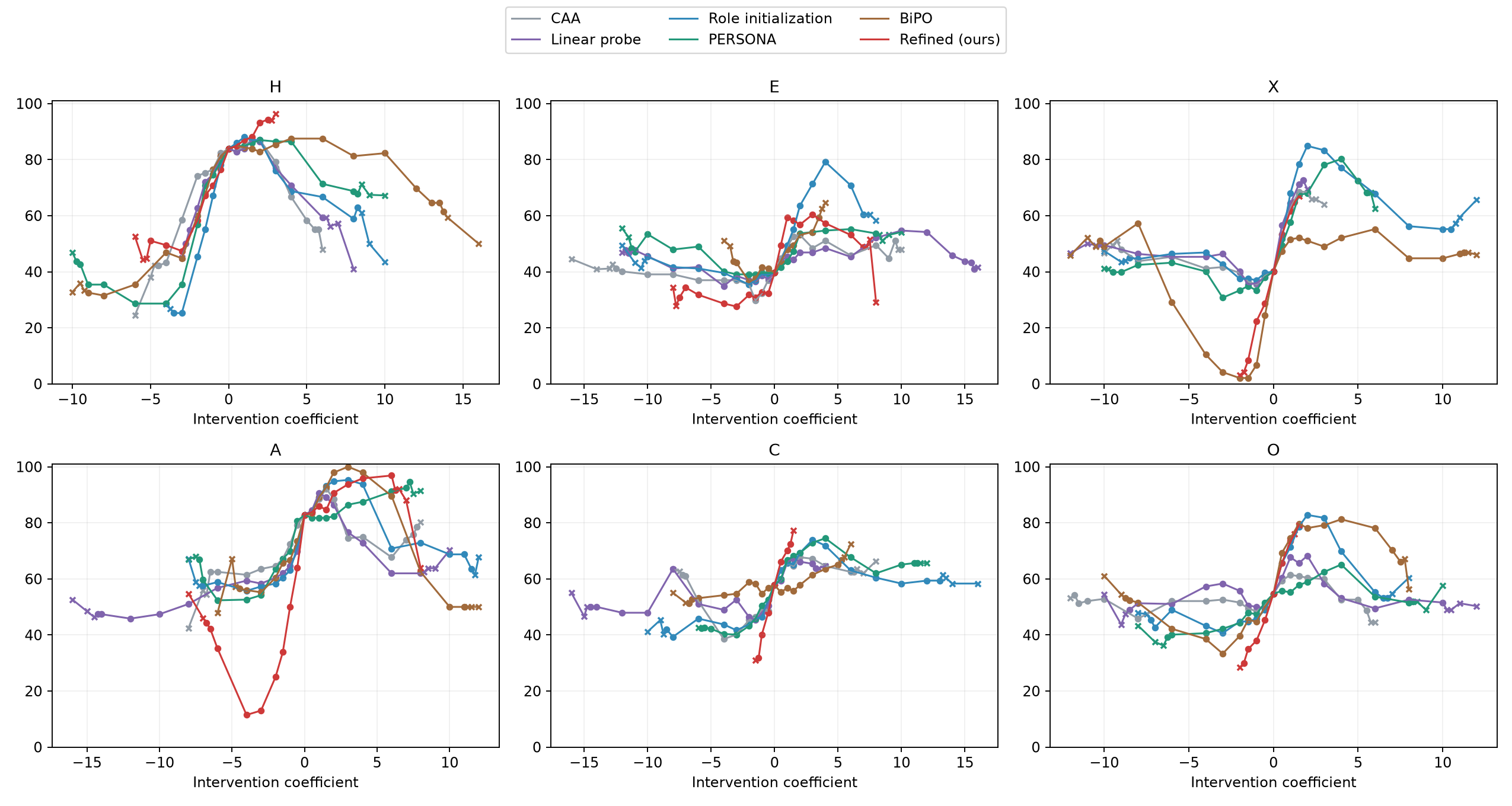}}
\caption*{Figure I1. Expanded complete-response score curves. Dots meet the initial 48-response validity threshold of at least 90\%; crosses do not.}
\end{figure}

\begin{figure}
\centering
\pandocbounded{\includegraphics[keepaspectratio,alt={Figure I2. Complete-response validity for the same conditions. Dashed lines mark the 90\% threshold.}]{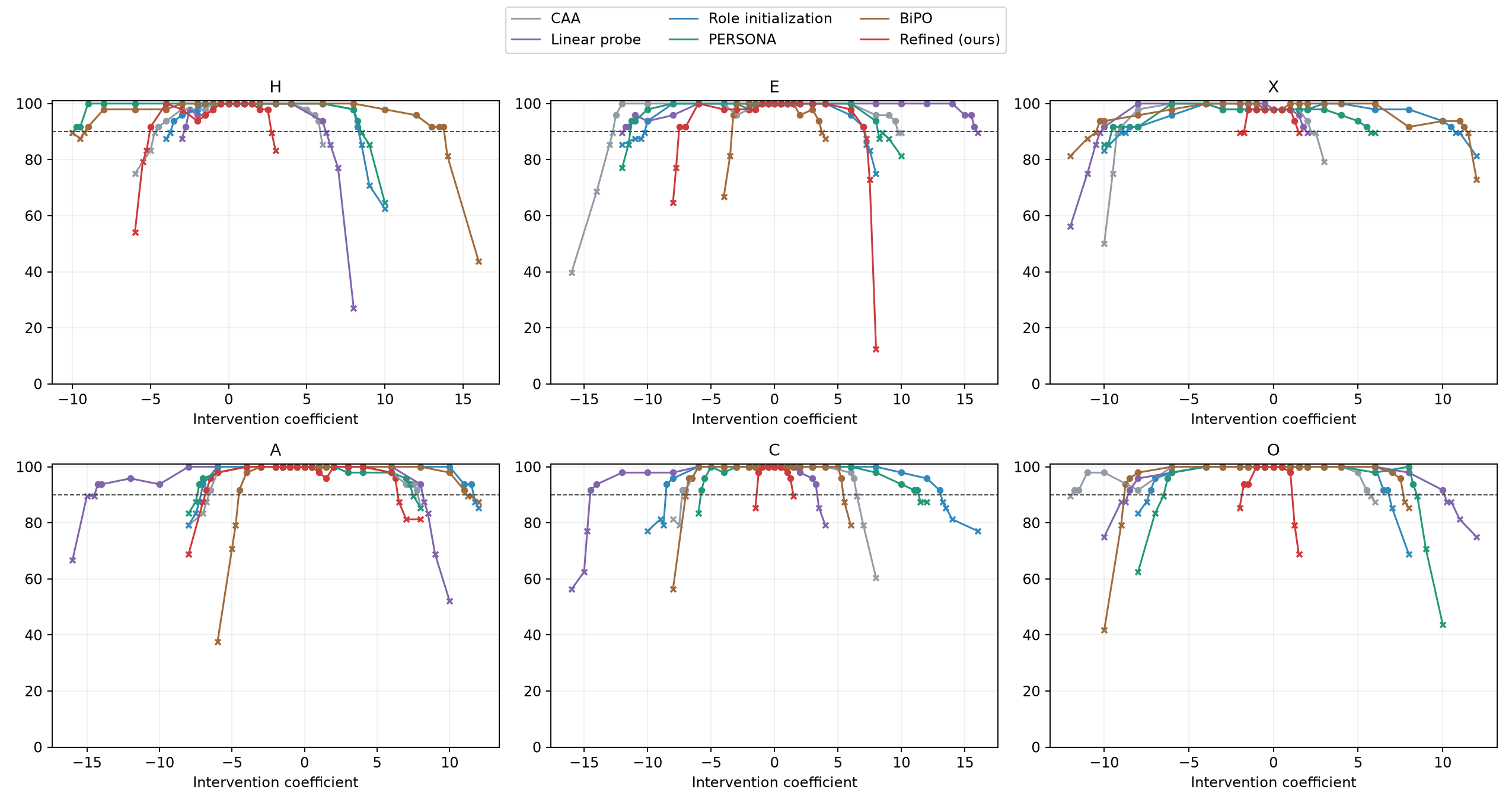}}
\caption*{Figure I2. Complete-response validity for the same conditions. Dashed lines mark the 90\% threshold.}
\end{figure}

\textbf{Table I1. Refined observed ranges.} Boundary columns give absolute-strength passing/failing brackets based on the first 48 answers, not confidence intervals.

{\small\begin{longtable}[]{@{}
  >{\raggedright\arraybackslash}p{(\linewidth - 8\tabcolsep) * \real{0.1875}}
  >{\raggedright\arraybackslash}p{(\linewidth - 8\tabcolsep) * \real{0.1875}}
  >{\raggedleft\arraybackslash}p{(\linewidth - 8\tabcolsep) * \real{0.2500}}
  >{\raggedright\arraybackslash}p{(\linewidth - 8\tabcolsep) * \real{0.1875}}
  >{\raggedright\arraybackslash}p{(\linewidth - 8\tabcolsep) * \real{0.1875}}@{}}
\toprule\noalign{}
\begin{minipage}[b]{\linewidth}\raggedright
Dimension
\end{minipage} & \begin{minipage}[b]{\linewidth}\raggedright
Minimum--maximum
\end{minipage} & \begin{minipage}[b]{\linewidth}\raggedleft
Span
\end{minipage} & \begin{minipage}[b]{\linewidth}\raggedright
Negative bracket
\end{minipage} & \begin{minipage}[b]{\linewidth}\raggedright
Positive bracket
\end{minipage} \\
\midrule\noalign{}
\endhead
\bottomrule\noalign{}
\endlastfoot
H & 47.40---94.27 & 46.87 & 5.00---5.25 & 2.50---2.75 \\
E & 27.60---60.42 & 32.81 & 7.50---7.75 & 7.00---7.25 \\
X & 8.33---64.84 & 56.51 & 1.50---1.75 & 1.25---1.50 \\
A & 11.46---96.88 & 85.42 & 6.75---7.00 & 6.25---6.50 \\
C & 31.77---72.40 & 40.62 & 1.25---1.50 & 1.25---1.50 \\
O & 29.95---73.44 & 43.49 & 1.75---2.00 & 1.00---1.25 \\
\end{longtable}}

\subsection*{I.3 Boundary Remeasurement and Records}\label{i.3-boundary-remeasurement-and-records}

Each boundary endpoint receives another 48 answers, pooled with the original 48. Of 144 endpoints, 44 change their pass/fail classification after pooling 96 responses. Thus 0.25 is search resolution, not boundary confidence precision. Excluding initially passing points that fail the pooled threshold gives mean usable spans of 50.78 for refinement, 45.14 for initialization, 41.23 for BiPO, 37.72 for adapted PERSONA, 29.47 for CAA, and 29.17 for the independent probe. Refinement retains the largest mean span.

The scan reuses matching strengths from 294 fixed-curve conditions and measures 636 further conditions, comprising 30,528 response attempts. All raw-record hashes were verified. Scanning does not retrain directions or reselect layers. Per-branch boundaries, norms, and pooled repeats are preserved in \nolinkurl{evidence/hexaco_six_methods_range_records.json}.

\subsection*{I.4 Six-Dimensional Joint Measurement}\label{i.4-six-dimensional-joint-measurement}

Each method uses −1, 0, and +1 per dimension, giving 729 combinations and 4,374 total nodes with 419,904 item measurements. All are complete. Models, LoRA, directions, layers, and full-input/decode injection match the single-axis experiments; PERSONA −1 still denotes the independent low pole. Each node evaluates 48 items under two numeric mappings at temperature 0.7. Conditional expectations over legal 1--5 labels are keyed to 0--100 and averaged equally over sixteen presentations per dimension.

Quality requires mean full-vocabulary legal numeric mass \ensuremath{\geq}0.98 per dimension and an absolute gap \ensuremath{\leq}25 between dimension means under the two mappings. With no behavioral A/B interface, only these two conditions apply, rather than the four-part cognitive--behavioral gate. A legal probability of 98\% is not a complete-generation validity rate. Complete-response curves and first-token catalogs use distinct scoring interfaces.

Seed 2026102606 generates 10,000 six-dimensional unit-cube targets, rescaled to each domain. Strict coverage requires one candidate with all coordinate errors \textless10 and total error \textless36. MAE averages coordinatewise absolute errors of ordinary nearest-L1 retrieval. The statistics describe the measured catalog, not a new independent generation repeat. Test items are held out from training and selection, while catalog construction and static coverage share the same evaluation interface.

\textbf{Table I2. Strict coverage and matching errors across target domains.} All results use the same quality conditions. Coverage is in percent and MAE in score points.

{\small\begin{longtable}[]{@{}
  >{\raggedright\arraybackslash}p{(\linewidth - 12\tabcolsep) * \real{0.1111}}
  >{\raggedleft\arraybackslash}p{(\linewidth - 12\tabcolsep) * \real{0.1481}}
  >{\raggedleft\arraybackslash}p{(\linewidth - 12\tabcolsep) * \real{0.1481}}
  >{\raggedleft\arraybackslash}p{(\linewidth - 12\tabcolsep) * \real{0.1481}}
  >{\raggedleft\arraybackslash}p{(\linewidth - 12\tabcolsep) * \real{0.1481}}
  >{\raggedleft\arraybackslash}p{(\linewidth - 12\tabcolsep) * \real{0.1481}}
  >{\raggedleft\arraybackslash}p{(\linewidth - 12\tabcolsep) * \real{0.1481}}@{}}
\toprule\noalign{}
\begin{minipage}[b]{\linewidth}\raggedright
Method
\end{minipage} & \begin{minipage}[b]{\linewidth}\raggedleft
25--75 strict
\end{minipage} & \begin{minipage}[b]{\linewidth}\raggedleft
25---75 MAE
\end{minipage} & \begin{minipage}[b]{\linewidth}\raggedleft
20--80 strict
\end{minipage} & \begin{minipage}[b]{\linewidth}\raggedleft
20---80 MAE
\end{minipage} & \begin{minipage}[b]{\linewidth}\raggedleft
30--70 strict
\end{minipage} & \begin{minipage}[b]{\linewidth}\raggedleft
30---70 MAE
\end{minipage} \\
\midrule\noalign{}
\endhead
\bottomrule\noalign{}
\endlastfoot
CAA & 4.94 & 10.20 & 2.61 & 12.14 & 10.20 & 8.37 \\
Independent probe & 4.23 & 10.81 & 2.22 & 12.72 & 7.47 & 9.02 \\
Role initialization & 9.08 & 8.70 & 4.60 & 10.62 & 20.80 & 6.87 \\
Adapted PERSONA & 5.31 & 10.04 & 2.97 & 11.98 & 10.08 & 8.23 \\
BiPO & 3.95 & 10.39 & 2.30 & 11.99 & 7.42 & 8.78 \\
\textbf{Refined} & \textbf{12.56} & \textbf{8.13} & \textbf{6.49} & \textbf{9.87} & \textbf{25.24} & \textbf{6.50} \\
\end{longtable}}

\textbf{Table I3. Unfiltered results in 25--75.} Other domains and bin statistics are retained in the supporting records.

{\small\begin{longtable}[]{@{}lrr@{}}
\toprule\noalign{}
Method & Strict coverage (\%) & Matching MAE \\
\midrule\noalign{}
\endhead
\bottomrule\noalign{}
\endlastfoot
CAA & 5.52 & 9.98 \\
Independent probe & 6.29 & 9.87 \\
Role initialization & 9.37 & 8.66 \\
Adapted PERSONA & 5.31 & 10.04 \\
BiPO & 4.45 & 10.21 \\
Refined & 13.48 & 8.07 \\
\end{longtable}}

The comparison fixes each method's number of combinations and item measurements, without assuming identical execution time across selected layers. Initial and refined methods each select layers on development data, so this experiment evaluates their final construction pipelines. Section 4.5 separately isolates refinement with layers fixed. Node-level results, pre/post-filter summaries, and independent raw-probability checks are indexed in Appendix H.4.

%% file: main.bbl
\begin{thebibliography}{50}
\bibitem[Anusic et~al.(2009)]{ref40}
Ivana Anusic, Ulrich Schimmack, Rebecca T. Pinkus, Penelope Lockwood. 2009. \href{https://pubmed.ncbi.nlm.nih.gov/19968424/}{The Nature and Structure of Correlations Among Big Five Ratings: The Halo-Alpha-Beta Model}. \emph{Journal of Personality and Social Psychology 97(6), 1142--1156. DOI: 10.1037/a0017159.}

\bibitem[Argyle et~al.(2023)]{ref2}
Lisa P. Argyle, Ethan C. Busby, Nancy Fulda, Joshua R. Gubler, Christopher Rytting, David Wingate. 2023. \href{https://doi.org/10.1017/pan.2023.2}{Out of One, Many: Using Language Models to Simulate Human Samples}. \emph{Political Analysis. DOI: 10.1017/pan.2023.2.}

\bibitem[Aron \& Aron(1997)]{ref38}
Elaine N. Aron, Arthur Aron. 1997. \href{https://www.hsperson.com/pdf/JPSP_Aron_and_Aron_97_Sensitivity_vs_I_and_N.pdf}{Sensory-processing sensitivity and its relation to introversion and emotionality}. \emph{Journal of Personality and Social Psychology, 73(2), 345--368. DOI: 10.1037/0022-3514.73.2.345.}

\bibitem[Bigelow et~al.(2026)]{ref45}
Eric Bigelow, Daniel Wurgaft, YingQiao Wang, Noah Goodman, Tomer Ullman, Hidenori Tanaka, Ekdeep Singh Lubana. 2026. \href{https://arxiv.org/html/2511.00617v2}{Belief Dynamics Reveal the Dual Nature of In-Context Learning and Activation Steering}. \emph{arXiv:2511.00617v2, revised 12 March 2026; first submitted November 2025. DOI: 10.48550/arXiv.2511.00617.}

\bibitem[Blais \& Weber(2006)]{ref34}
Ann-Renée Blais, Elke U. Weber. 2006. \href{https://www.cambridge.org/core/journals/judgment-and-decision-making/article/domainspecific-risktaking-dospert-scale-for-adultpopulations/419BDAAF215313A7EE216F51F38AD205}{A Domain-Specific Risk-Taking (DOSPERT) scale for adult populations}. \emph{Judgment and Decision Making, 1(1), 33--47. DOI: 10.1017/S1930297500000334.}

\bibitem[Bungartz \& Griebel(2004)]{ref29}
Hans-Joachim Bungartz, Michael Griebel. 2004. \href{https://www.cambridge.org/core/journals/acta-numerica/article/abs/sparse-grids/47EA2993DB84C9D231BB96ECB26F615C}{Sparse grids}. \emph{Acta Numerica 13, 147--269. DOI: 10.1017/S0962492904000182.}

\bibitem[Cao et~al.(2024)]{ref47}
Yuanpu Cao, Tianrong Zhang, Bochuan Cao, Ziyi Yin, Lu Lin, Fenglong Ma, Jinghui Chen. 2024. \href{https://proceedings.neurips.cc/paper_files/paper/2024/hash/58cbe393b4254da8966780a40d023c0b-Abstract-Conference.html}{Personalized Steering of Large Language Models: Versatile Steering Vectors Through Bi-directional Preference Optimization}. \emph{Advances in Neural Information Processing Systems 37. DOI: 10.52202/079017-1567.}

\bibitem[Dathathri et~al.(2019)]{ref14}
Sumanth Dathathri, Andrea Madotto, Janice Lan, Jane Hung, Eric Frank, Piero Molino, Jason Yosinski, Rosanne Liu. 2019. \href{https://arxiv.org/abs/1912.02164}{Plug and Play Language Models: A Simple Approach to Controlled Text Generation}. \emph{arXiv:1912.02164.}

\bibitem[DeYoung(2006)]{ref39}
Colin G. DeYoung. 2006. \href{https://pubmed.ncbi.nlm.nih.gov/17144770/}{Higher-Order Factors of the Big Five in a Multi-Informant Sample}. \emph{Journal of Personality and Social Psychology 91(6), 1138--1151. DOI: 10.1037/0022-3514.91.6.1138.}

\bibitem[Du et~al.(2026)]{ref5}
Chengyu Du, Xintao Wang, Aili Chen, Weiyuan Li, Rui Xu, Junteng Liu, Zishan Huang, Rong Tian, Zijun Sun, Yuhao Li, Liheng Feng, Deming Ding, Pengyu Zhao, Yanghua Xiao. 2026. \href{https://aclanthology.org/2026.findings-acl.1283/}{HER: Human-like Reasoning and Reinforcement Learning for LLM Role-playing}. \emph{Findings of ACL 2026, 25725--25762. DOI: 10.18653/v1/2026.findings-acl.1283.}

\bibitem[Feng et~al.(2026)]{ref8}
Xiachong Feng, Liang Zhao, Weihong Zhong, Yichong Huang, Yuxuan Gu, Lingpeng Kong, Xiaocheng Feng, Bing Qin. 2026. \href{https://proceedings.iclr.cc/paper_files/paper/2026/hash/f66ce276436c4cd151535b03841ac713-Abstract-Conference.html}{PERSONA: Dynamic and Compositional Inference-Time Personality Control via Activation Vector Algebra}. \emph{International Conference on Learning Representations 2026.}

\bibitem[Føllesdal \& Soto(2022)]{ref41}
Hallvard Føllesdal, Christopher J. Soto. 2022. \href{https://www.colby.edu/wp-content/uploads/2022/05/Follesdal_Soto_2022.pdf}{The Norwegian Adaptation of the Big Five Inventory-2}. \emph{Frontiers in Psychology 13:858920. DOI: 10.3389/fpsyg.2022.858920.}

\bibitem[Ghasemi et~al.(2026)]{ref9}
Narges Ghasemi, Amir Ziashahabi, Salman Avestimehr, Jonathan May. 2026. \href{https://arxiv.org/abs/2606.22357v2}{ORBIT: Training-Free Multi-Attribute Behavioral Steering via Orthogonal Subspace Rotation}. \emph{arXiv:2606.22357v2.}

\bibitem[Hastie \& Tibshirani(1986)]{ref44}
Trevor Hastie, Robert Tibshirani. 1986. \href{https://doi.org/10.1214/ss/1177013604}{Generalized Additive Models}. \emph{Statistical Science 1(3), 297--310. DOI: 10.1214/ss/1177013604.}

\bibitem[Hoppe et~al.(2026)]{ref7}
Florian Hoppe, David Khachaturov, Robert D. Mullins, Mark Huasong Meng. 2026. \href{https://proceedings.mlr.press/v306/hoppe26a.html}{Controllable and Explainable Personality Sliders for LLMs at Inference Time}. \emph{Proceedings of the 43rd International Conference on Machine Learning, Proceedings of Machine Learning Research 306, 43871--43895.}

\bibitem[Hu et~al.(2021)]{ref10}
Edward J. Hu, Yelong Shen, Phillip Wallis, Zeyuan Allen-Zhu, Yuanzhi Li, Shean Wang, Lu Wang, Weizhu Chen. 2021. \href{https://arxiv.org/abs/2106.09685}{LoRA: Low-Rank Adaptation of Large Language Models}. \emph{arXiv:2106.09685.}

\bibitem[Huang et~al.(2024)]{ref30}
Chengsong Huang, Qian Liu, Bill Yuchen Lin, Tianyu Pang, Chao Du, Min Lin. 2024. \href{https://arxiv.org/abs/2307.13269v3}{LoraHub: Efficient Cross-Task Generalization via Dynamic LoRA Composition}. \emph{arXiv:2307.13269v3; COLM 2024. DOI: 10.48550/arXiv.2307.13269.}

\bibitem[International Personality Item Pool(2026)]{ref32}
International Personality Item Pool. 2026. \href{https://ipip.ori.org/30FacetNEO-PI-RItems.htm}{IPIP items and scoring keys for the 30 facets of the NEO-PI-R; Johnson 120-item selection}. \emph{Official public-domain item and scoring-key resource; accessed 30 September 2026.}

\bibitem[Jin et~al.(2026)]{ref43}
Zehao Jin, Junran Wang, Ruixuan Deng, Jiahao Chen, Jingyuan Zhang, Yuxuan Zhang, Xinjie Shen. 2026. \href{https://arxiv.org/html/2609.36388v1}{Persona Dosing: Calibrated Activation Steering for Graded Trait Control}. \emph{arXiv:2609.36388v1, 28 September 2026. DOI: 10.48550/arXiv.2609.36388.}

\bibitem[Johnson(2014)]{ref33}
John A. Johnson. 2014. \href{https://doi.org/10.1016/j.jrp.2014.05.003}{Measuring thirty facets of the Five Factor Model with a 120-item public domain inventory: Development of the IPIP-NEO-120}. \emph{Journal of Research in Personality 51, 78--89. DOI: 10.1016/j.jrp.2014.05.003.}

\bibitem[Li et~al.(2026)]{ref25}
Chang-Jin Li, Jiyuan Zhang, Yun Tang, Jian Li. 2026. \href{https://doi.org/10.1016/j.chbr.2026.100964}{Automatic item generation for personality situational judgment tests with large language models}. \emph{Computers in Human Behavior Reports, 21, Article 100964. DOI: 10.1016/j.chbr.2026.100964.}

\bibitem[Li et~al.(2023)]{ref46}
Kenneth Li, Oam Patel, Fernanda Viégas, Hanspeter Pfister, Martin Wattenberg. 2023. \href{https://papers.nips.cc/paper_files/paper/2023/hash/81b8390039b7302c909cb769f8b6cd93-Abstract-Conference.html}{Inference-Time Intervention: Eliciting Truthful Answers from a Language Model}. \emph{Advances in Neural Information Processing Systems 36. DOI: 10.52202/075280-1797.}

\bibitem[Lu et~al.(2026)]{ref6}
Christina Lu, Jack Gallagher, Jonathan Michala, Kyle Fish, Jack Lindsey. 2026. \href{https://arxiv.org/abs/2601.10387}{The Assistant Axis: Situating and Stabilizing the Default Persona of Language Models}. \emph{arXiv:2601.10387.}

\bibitem[Moon et~al.(2024)]{ref11}
Suhong Moon, Marwa Abdulhai, Minwoo Kang, Joseph Suh, Widyadewi Soedarmadji, Eran Kohen Behar, David M. Chan. 2024. \href{https://aclanthology.org/2024.emnlp-main.1110/}{Virtual Personas for Language Models via an Anthology of Backstories}. \emph{Proceedings of EMNLP 2024. DOI: 10.18653/v1/2024.emnlp-main.1110.}

\bibitem[Mouret \& Clune(2015)]{ref27}
Jean-Baptiste Mouret, Jeff Clune. 2015. \href{https://arxiv.org/abs/1504.04909}{Illuminating search spaces by mapping elites}. \emph{arXiv:1504.04909.}

\bibitem[Murphy et~al.(2011)]{ref36}
Ryan O. Murphy, Kurt A. Ackermann, Michel J. J. Handgraaf. 2011. \href{https://www.cambridge.org/core/journals/judgment-and-decision-making/article/measuring-social-value-orientation/78981D731BFB89AFCFC789D40FD8C11F}{Measuring Social Value Orientation}. \emph{Judgment and Decision Making, 6(8), 771--781. DOI: 10.1017/S1930297500004204.}

\bibitem[Nguyen et~al.(2026)]{ref20}
Dung Viet Nguyen, Yen Pham, Hieu Vu, Lei Zhang, Tan Nguyen. 2026. \href{https://proceedings.iclr.cc/paper_files/paper/2026/hash/fa5617c176e76fee83f3f9947fdf9f3f-Abstract-Conference.html}{Activation Steering with a Feedback Controller}. \emph{International Conference on Learning Representations 2026.}

\bibitem[Nguyen et~al.(2025)]{ref48}
Duy Nguyen, Archiki Prasad, Elias Stengel-Eskin, Mohit Bansal. 2025. \href{https://aclanthology.org/2025.acl-long.1007/}{Multi-Attribute Steering of Language Models via Targeted Intervention}. \emph{Proceedings of ACL 2025, Volume 1: Long Papers, 20619--20634. DOI: 10.18653/v1/2025.acl-long.1007.}

\bibitem[Ni et~al.(2026)]{ref1}
Bo Ni, Yu Wang, Leyao Wang, et al.~2026. \href{https://aclanthology.org/2026.eacl-long.200/}{A Survey on LLM-based Conversational User Simulation}. \emph{Proceedings of EACL 2026, Volume 1: Long Papers, 4266--4301. DOI: 10.18653/v1/2026.eacl-long.200.}

\bibitem[Oozeer et~al.(2025)]{ref19}
Narmeen Fatimah Oozeer, Luke Marks, Fazl Barez, Amir Abdullah. 2025. \href{https://aclanthology.org/2025.findings-emnlp.1278/}{Beyond Linear Steering: Unified Multi-Attribute Control for Language Models}. \emph{Findings of EMNLP 2025, 23513--23557. DOI: 10.18653/v1/2025.findings-emnlp.1278.}

\bibitem[Park et~al.(2023)]{ref3}
Joon Sung Park, Joseph C. O'Brien, Carrie J. Cai, Meredith Ringel Morris, Percy Liang, Michael S. Bernstein. 2023. \href{https://arxiv.org/abs/2304.03442}{Generative Agents: Interactive Simulacra of Human Behavior}. \emph{arXiv:2304.03442.}

\bibitem[Park et~al.(2026)]{ref12}
Joon Sung Park, Carolyn Q. Zou, Jonne Kamphorst, Niles Egan, Aaron Shaw, Benjamin Mako Hill, Carrie Cai, Meredith Ringel Morris, Percy Liang, Robb Willer, Michael S. Bernstein. 2026. \href{https://arxiv.org/abs/2411.10109v3}{LLM Agents Grounded in Self-Reports Enable General-Purpose Simulation of Individuals}. \emph{arXiv:2411.10109v3.}

\bibitem[Pugh et~al.(2016)]{ref28}
Justin K. Pugh, L. B. Soros, Kenneth O. Stanley. 2016. \href{https://www.frontiersin.org/journals/robotics-and-ai/articles/10.3389/frobt.2016.00040/full}{Quality Diversity: A New Frontier for Evolutionary Computation}. \emph{Frontiers in Robotics and AI 3:40. DOI: 10.3389/frobt.2016.00040.}

\bibitem[Qwen Team(2026)]{ref31}
Qwen Team. 2026. \href{https://huggingface.co/Qwen/Qwen3.5-27B}{Qwen3.5-27B: Official Model Card and Configuration}. \emph{Hugging Face model repository; accessed 13 October 2026.}

\bibitem[Richins \& Dawson(1992)]{ref37}
Marsha L. Richins, Scott Dawson. 1992. \href{https://academic.oup.com/jcr/article-abstract/19/3/303/1786697}{A Consumer Values Orientation for Materialism and Its Measurement: Scale Development and Validation}. \emph{Journal of Consumer Research, 19(3), 303--316. DOI: 10.1086/209304.}

\bibitem[Rimsky et~al.(2024)]{ref16}
Nina Rimsky, Nick Gabrieli, Julian Schulz, Meg Tong, Evan Hubinger, Alexander Turner. 2024. \href{https://aclanthology.org/2024.acl-long.828/}{Steering Llama 2 via Contrastive Activation Addition}. \emph{Proceedings of ACL 2024, Volume 1: Long Papers, 15504--15522. DOI: 10.18653/v1/2024.acl-long.828.}

\bibitem[Scalena et~al.(2024)]{ref18}
Daniel Scalena, Gabriele Sarti, Malvina Nissim. 2024. \href{https://aclanthology.org/2024.blackboxnlp-1.34/}{Multi-property Steering of Large Language Models with Dynamic Activation Composition}. \emph{Proceedings of the 7th BlackboxNLP Workshop, 577--603. DOI: 10.18653/v1/2024.blackboxnlp-1.34.}

\bibitem[Serapio-García et~al.(2025)]{ref24}
Gregory Serapio-García, Mustafa Safdari, Clément Crepy, Luning Sun, Stephen Fitz, Peter Romero, Marwa Abdulhai, Aleksandra Faust, Maja Matarić. 2025. \href{https://www.nature.com/articles/s42256-025-01115-6}{A psychometric framework for evaluating and shaping personality traits in large language models}. \emph{Nature Machine Intelligence 7, 1954--1968. DOI: 10.1038/s42256-025-01115-6.}

\bibitem[Skifstad et~al.(2026)]{ref21}
Julian Skifstad, Xinyue Annie Yang, Glen Chou. 2026. \href{https://proceedings.mlr.press/v306/skifstad26a.html}{Local Linearity of LLMs Enables Activation Steering via Model-Based Linear Optimal Control}. \emph{Proceedings of ICML 2026, Proceedings of Machine Learning Research 306, 113885--113919.}

\bibitem[Strathman et~al.(1994)]{ref35}
Alan Strathman, Faith Gleicher, David S. Boninger, C. S. Edwards. 1994. \href{https://doi.org/10.1037/0022-3514.66.4.742}{The consideration of future consequences: Weighing immediate and distant outcomes of behavior}. \emph{Journal of Personality and Social Psychology, 66(4), 742--752. DOI: 10.1037/0022-3514.66.4.742.}

\bibitem[Taimeskhanov et~al.(2026)]{ref22}
Magamed Taimeskhanov, Samuel Vaiter, Damien Garreau. 2026. \href{https://proceedings.mlr.press/v306/taimeskhanov26a.html}{Towards Understanding Steering Strength}. \emph{Proceedings of ICML 2026, Proceedings of Machine Learning Research 306, 118084--118133.}

\bibitem[Turpin et~al.(2023)]{ref26}
Miles Turpin, Julian Michael, Ethan Perez, Samuel R. Bowman. 2023. \href{https://arxiv.org/abs/2305.04388}{Language Models Don't Always Say What They Think: Unfaithful Explanations in Chain-of-Thought Prompting}. \emph{arXiv:2305.04388.}

\bibitem[Wang et~al.(2024)]{ref13}
Noah Wang, Z. Y. Peng, Haoran Que, et al.~2024. \href{https://aclanthology.org/2024.findings-acl.878/}{RoleLLM: Benchmarking, Eliciting, and Enhancing Role-Playing Abilities of Large Language Models}. \emph{Findings of ACL 2024, 14743--14777. DOI: 10.18653/v1/2024.findings-acl.878.}

\bibitem[Wang et~al.(2025)]{ref50}
Kuang Wang, Xianfei Li, Shenghao Yang, Li Zhou, Feng Jiang, Haizhou Li. 2025. \href{https://aclanthology.org/2025.acl-long.1025/}{Know You First and Be You Better: Modeling Human-Like User Simulators via Implicit Profiles}. \emph{Proceedings of ACL 2025, Volume 1: Long Papers, 21082--21107. DOI: 10.18653/v1/2025.acl-long.1025.}

\bibitem[Wu et~al.(2026)]{ref4}
Xiaoting Wu, Yi Huang, Chunyang Gao, Mengfei Guo, Jingyu Yao, Junlan Feng. 2026. \href{https://aclanthology.org/2026.acl-long.1989/}{Thinking Alignment of Scenario-Oriented User Simulation}. \emph{Proceedings of ACL 2026, Volume 1: Long Papers, 42932--42945. DOI: 10.18653/v1/2026.acl-long.1989.}

\bibitem[Wu et~al.(2024)]{ref17}
Zhengxuan Wu, Aryaman Arora, Zheng Wang, Atticus Geiger, Dan Jurafsky, Christopher D. Manning, Christopher Potts. 2024. \href{https://proceedings.neurips.cc/paper_files/paper/2024/hash/75008a0fba53bf13b0bb3b7bff986e0e-Abstract-Conference.html}{ReFT: Representation Finetuning for Language Models}. \emph{Advances in Neural Information Processing Systems 37. DOI: 10.52202/079017-2041.}

\bibitem[Xu et~al.(2026)]{ref23}
Ziwen Xu, Chenyan Wu, Hengyu Sun, Haiwen Hong, Mengru Wang, Yunzhi Yao, Longtao Huang, Hui Xue, Shumin Deng, Zhixuan Chu, Huajun Chen, Ningyu Zhang. 2026. \href{https://aclanthology.org/2026.acl-long.1463/}{Why Steering Works: Toward a Unified View of Language Model Parameter Dynamics}. \emph{Proceedings of ACL 2026, Volume 1: Long Papers, 31719--31736. DOI: 10.18653/v1/2026.acl-long.1463.}

\bibitem[Yang \& Klein(2021)]{ref15}
Kevin Yang, Dan Klein. 2021. \href{https://aclanthology.org/2021.naacl-main.276/}{FUDGE: Controlled Text Generation With Future Discriminators}. \emph{Proceedings of NAACL-HLT 2021, 3511--3535. DOI: 10.18653/v1/2021.naacl-main.276.}

\bibitem[Yang et~al.(2026)]{ref49}
Yutong Yang, Chenxi Miao, Weikang Li, Yunfang Wu. 2026. \href{https://arxiv.org/abs/2605.29791v2}{ActTraitBench: Quantifying the Knowledge-Decision Gap in Large Language Models via Human-Grounded Behavioral Validation}. \emph{arXiv:2605.29791v2; accepted to Findings of EMNLP 2026. DOI: 10.48550/arXiv.2605.29791.}

\bibitem[Zhang et~al.(2026)]{ref42}
Rongzhi Zhang, Liqin Ye, Yuzhao Heng, Xiang Chen, Tong Yu, Lingkai Kong, Sudheer Chava, Chao Zhang. 2026. \href{https://arxiv.org/html/2510.12121v2}{Precise Attribute Intensity Control in Large Language Models via Targeted Representation Editing}. \emph{arXiv:2510.12121v2, revised 18 February 2026; first submitted 14 October 2025. DOI: 10.48550/arXiv.2510.12121.}

\end{thebibliography}
